# Haze Visibility Enhancement: A Survey and Quantitative Benchmarking

Yu Li, Shaodi You, Michael S. Brown, and Robby T. Tan

***Abstract*—This paper provides a comprehensive survey of methods dealing with visibility enhancement of images taken in hazy or foggy scenes. The survey begins with discussing the optical models of atmospheric scattering media and image formation. This is followed by a survey of existing methods, which are grouped to multiple image methods, polarizing filters based methods, methods with known depth, and single-image methods. We also provide a benchmark of a number of well known single-image methods, based on a recent dataset provided by Fattal [1] and our newly generated scattering media dataset that contains ground truth images for quantitative evaluation. To our knowledge, this is the first benchmark using numerical metrics to evaluate dehazing techniques. This benchmark allows us to objectively compare the results of existing methods and to better identify the strengths and limitations of each method.**

## I. Introduction

FOG and haze are two of the most common real world phenomena caused by atmospheric particles. Images captured in foggy and hazy scenes suffer from noticeable degradation of contrast and visibility (Figure 1). Many computer vision and image processing algorithms suffer from the visibility degradation, since most of them assume clear scenes under good weather. Addressing this problem via so-called "dehazing" or "defogging" algorithms is therefore of practical importance.

The degradation in hazy and foggy images can be physically attributed to floating particles in the atmosphere that absorb and scatter light in the environment [2]. This scattering and absorption reduces the direct transmission from the scene to the camera and adds another layer of the surrounding scattered light, known as airlight [3]. The attenuated direct transmission causes the intensity from the scene to be weaker, while the airlight causes the appearance of the scene to be washed out.

In the past two decades, there has been significant progress in methods that use images taken in hazy scenes. Early work by Cozman and Krotkov [4] and Nayar and Narasimhan [5], [6] use atmospheric cues to estimate depth. Since then, a number of methods have been introduced to explicitly enhance visibility, which can be grouped into four categories: multi-image based methods (*e.g.* [7], [8], [9]), polarizing filter based methods (*e.g.* [10], [11]), methods using known depth or geometrical information (*e.g.* [12], [13], [14]) and single image methods (*e.g.* [15], [16], [1], [17], [18]).

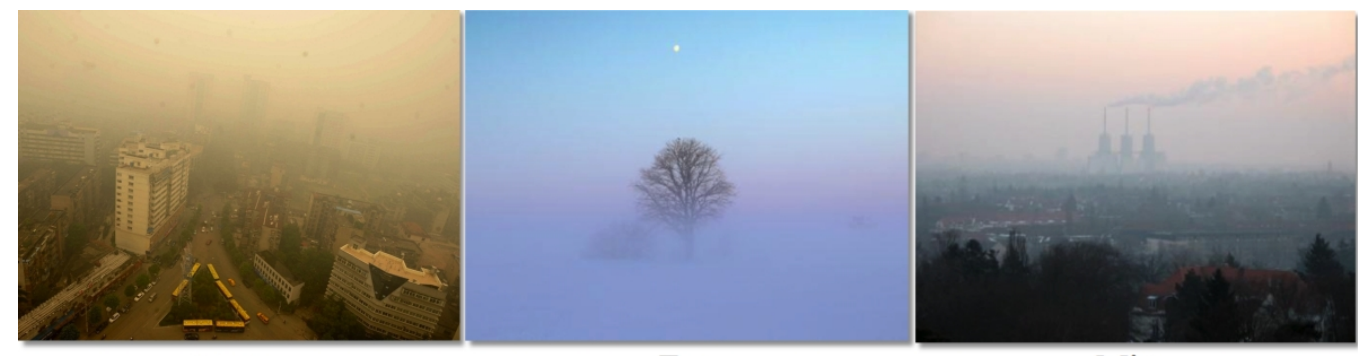

Haze Fog Mist

Fig. 1. Several examples of images showing the visual phenomena of atmospheric particles. Most of them exhibit significant visibility degradation.

The first contribution of this paper is to provide a detailed survey on methods focusing on scattering-media visibility enhancement. Our survey provides a holistic view on most of the existing methods. Starting with a brief introduction of the atmospheric scattering optics in Section II, in Section III we provide a chronological survey of visibility enhancement methods in atmospheric scattering media. Particular emphasize is placed on the last category of single-image methods as they offer the most flexibility by not requiring additional information.

As part of this survey, we also provide a quantitative benchmarking of a number of the single-image methods. Obtaining quantitative results is challenging as it is difficult to capture ground truth examples for where the same scene has been imaged with and without scattering particles. The work by Fattal [1] synthesized a dataset of by using natural images which associate depth maps that can be used to simulate the spatially varying attenuating in haze and fog images. We have generated an additional dataset using a physically-based rendering to simulate environments with scattered particles. Section IV provides the results of the different methods using both on Fattal's dataset [1] and also our newly generated benchmark dataset.

Our paper is concluded in Section V with a discussion on the current state of image dehazing methods and the findings from the benchmark results. In particular, we discuss current limitations with existing methods and possible avenues for research for future methods.

## II. Atmospheric Scattering Model

Haze is a common atmospheric phenomenon resulting from air pollution such as dust, smoke and other dry particles that obscure the clarity of the sky. Sources for haze particles include farming, traffic, industry, and wildfire. As listed in Table I, the particle size varies from $10^{-2} - 1\mu m$ and the

Y. Li is with the Advanced Digital Sciences Center, Singapore (e-mail: li.yu@adsc.com.sg)

S. You is with Data61, Australia and Australian National University, Australia (e-mail: shaodi.you@anu.edu.au)

R. T. Tan is with Yale-NUS College and the National University of Singapore, Singapore (email: robby@gmail.com)

M. S. Brown is with York University, Canada (e-mail: mbrown@cse.yorku.ca)

TABLE I
WEATHER CONDITION AND THE PARTICLE TYPE, SIZE AND DENSITY.

| Weather | Particle type | Particle radius ($\mu m$) | Density ($cm^{-3}$) |
|---|---|---|---|
| Clean air | Molecule | $10^{-4}$ | $10^{19}$ |
| Haze | Aerosol | $10^{-2} - 1$ | $10 - 10^{3}$ |
| Fog | Water droplets | $1 - 10$ | $10 - 100$ |

We follow the definition of Hidy [19].

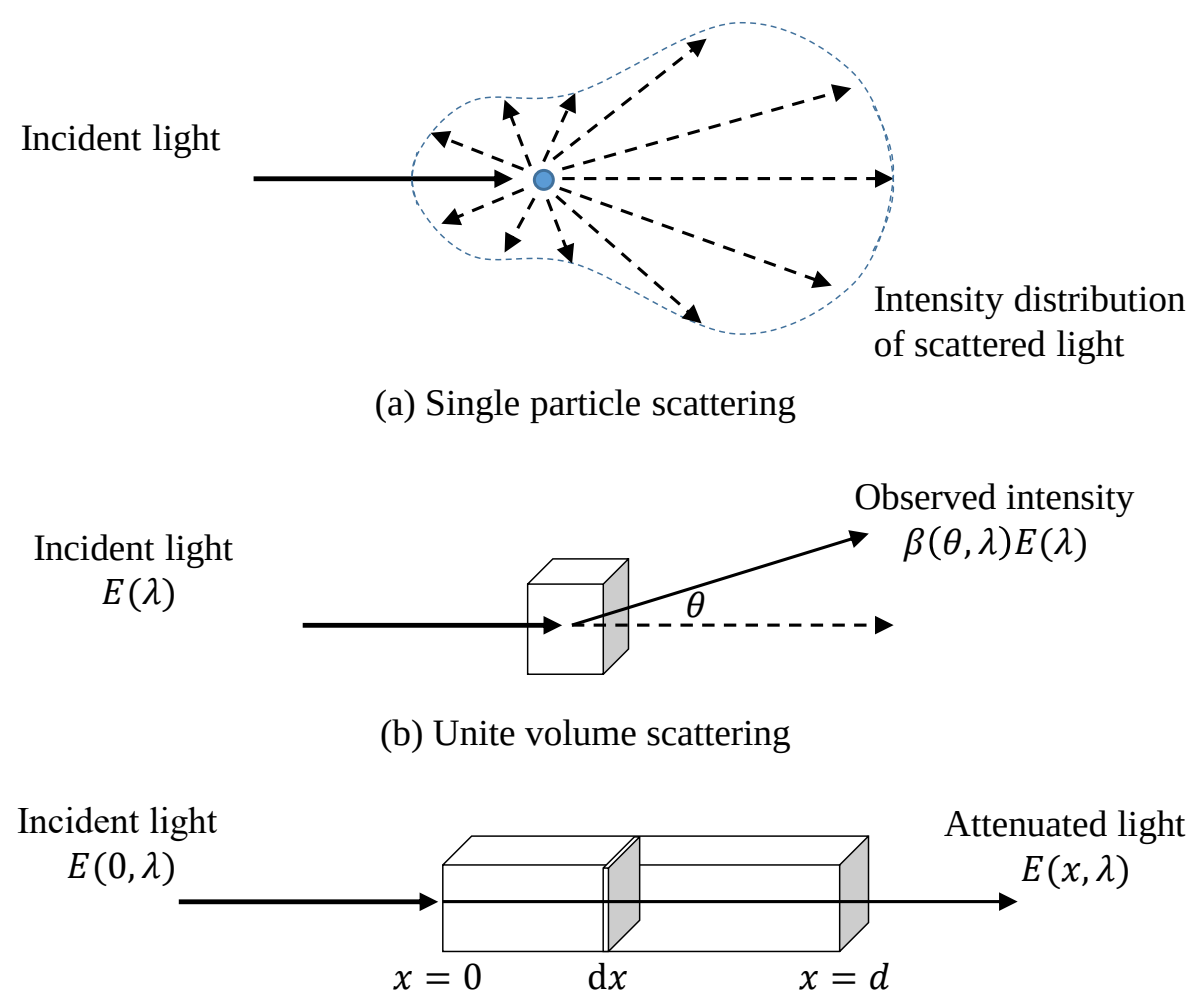

Fig. 2. (a) Single particle scattering; (b) Unit volume scattering and (c) Light attenuation over distance raised by scattering.

density varies from $10 - 10^3$ per $cm^3$. The particles cause visibility degradation and also color shift. Depending on the view-angle with respect to the sun and the types of the particles, haze may appear brownish or yellowish [20].

Unlike haze, fog (or mist) is caused by water droplets and/or ice crystals suspended in the air close to the earth surface [21]. As listed in Table I, the particle size varies from $1 - 10\mu m$ and the density varies from $10 - 100$ per $cm^3$. Generally, fog particles do not have their own color, and thus their color appearance depends mostly on the surrounding light colors.

## *A. Optical Modeling*

As illustrated in Figure 2(a), when a ray of light hits a particle, the particle will scatter the light to all directions with magnitudes depending on particle size, shape and incident light wavelengths. Since the directions of scattered rays are moving away from the particle, they are known as outbound rays or out-scattering rays. Accordingly, they are rays from all directions that hit a particle, and this is known as inbound rays or in-scattering rays. As well exploited by Minnaert [22], for a given particle type and incident light wavelength, the outbound light intensity can be modeled as a function between the angle of inbound and outbound light. In this paper, we are more interested in the statistical properties over a large number of particles. Thus, considering the particle density (Table I), and that each particle can be considered as an independent particle, we can have the statistical relationship between inbound light intensity $E$ and outbound light intensity $I$ [3]):

$$I(\theta, \lambda) = \beta_{p,\mathbf{x}}(\theta, \lambda)E(\lambda), \tag{1}$$

where $\beta_{p,\mathbf{x}}(\theta, \lambda)$ is called the *angular scattering coefficient*. The subindices of $\beta$, with $p$ indicating its dependency on particle type and density, and $\mathbf{x}$ indicates the dependency spatially. By integrating Eq. (1) over all spherical directions, we obtain the *total scattering coefficient*:

$$I(\lambda) = \beta_{p,\mathbf{x}}(\theta)E(\lambda). \tag{2}$$

**Direct Transmission** If we assume a particle medium consists of a small chunk with thickness $dx$, and a parallel light ray passes through every sheet, as illustrated in Figure 2(c) , then the change in irradiance at location $x$ is expressed as:

$$\frac{dE(x, \lambda)}{E(x, \lambda)} = -\beta_{p,x}(\lambda)dx. \tag{3}$$

Integrating this equation between $x = 0$ and $x = d$ gives us: $E(d, \lambda) = E_0(\lambda)e^{-\beta(\lambda)d}$ , where $E_0$ is the irradiance. The formula is known as the Beer-Lambert law.

For non-parallel rays of light, which occur more commonly for outdoor light, factoring in the inverse square law the equation becomes:

$$E(d, \lambda) = \frac{I_0(\lambda)e^{-\beta(\lambda)d}}{d^2} \tag{4}$$

where $I_0$ is the intensity of the source, assumed to be a point [3]. Moreover, as mentioned in [6], for overcast sky illumination, the last equation can be written as:

$$E(d, \lambda) = \frac{gL_\infty(\lambda)\rho(\lambda)e^{-\beta(\lambda)d}}{d^2}, \tag{5}$$

where $L_\infty$ is the light intensity at the horizon, $\rho$ is the reflectance of a scene point, and $g$ is the camera gain (assuming the light has been captured by a camera).

**Airlight** As illustrated in Fig. 3.a, besides light from a source (or reflected by objects) that pass through the medium and are transmitted towards the camera, there is environmental illumination in the atmosphere scattered by the same particles also towards the camera. The environmental illumination can be generated by direct sunlight, diffuse skylight, light reflected from the ground, and so on. This type of scattered light captured in the observer's cone of vision is called airlight [3].

Denote the light source as $I(x, \lambda)$, follow the unit volume scattering equation (Eq. (2) and Eq. (3)), we have:

$$\mathrm{d}I(x, \lambda) = \mathrm{d}V k\beta_{p,\mathbf{x}}(\lambda), \tag{6}$$

where $\mathrm{d}V = \mathrm{d}\omega x^2$ is a unit volume in the perspective cone. $k\beta_{p,\mathbf{x}}(\lambda)$ is the total scattering coefficient. $k$ is a constant representing the environmental illumination along the camera's line of sight. As with the mechanism for direct transmission in Eq.(4), this light source $dI$ passes through a small chunk of particles, and the outgoing light is expressed as:

$$dE(x, \lambda) = \frac{dI(x, \lambda)e^{-\beta(\lambda)x}}{x^2}, \tag{7}$$

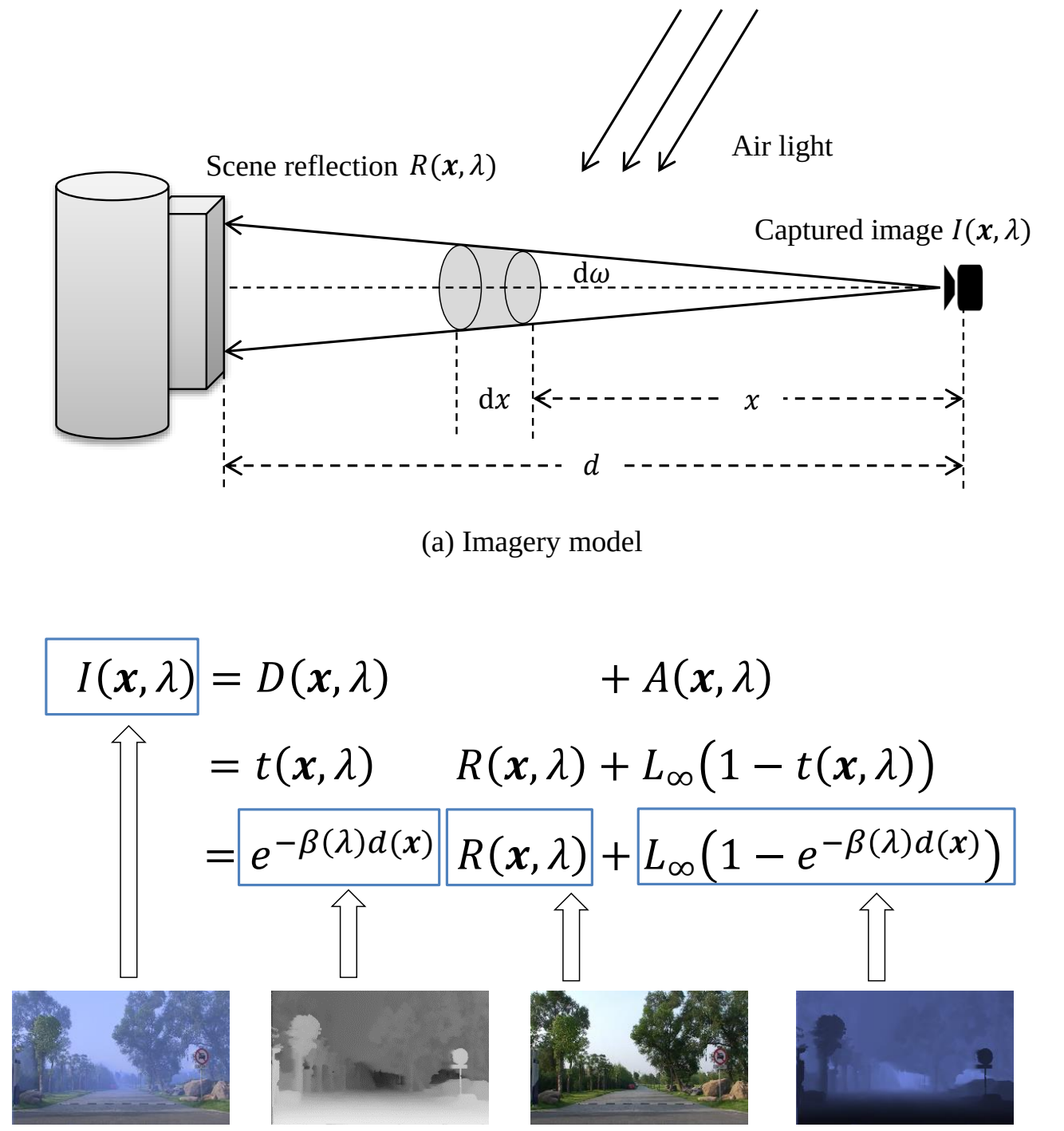

(a) Imagery model

(b) Formula of illumination components

Fig. 3. Visibility degradation problem in computer vision and computational imaging. (a) Imagery model: with the existence of atmospheric scattering media, light captured by a perspective camera have two components: one is the scene reflection attenuated by the scattering media (direct transmission), the other is the air-light (sunlight, diffused skylight and diffused ground light) scattered by media. (b) Formula and visual example of illumination components. Images are from [23].

where $x^2$ is due to the inverse square law of non-parallel rays of light. Therefore, the total radiance at distance $d$ from the camera can be obtained by integrating $dL = \frac{dE}{d\omega}$:

$$L(d, \lambda) = L_\infty \left(1 - e^{-\beta(\lambda)d}\right), \tag{8}$$

where $L_\infty = \sigma$, since when $d = \infty$, namely for an object at the horizon, then $L(d = \infty, \lambda) = \sigma$.

**Image Formation** As illustrated in Fig. 3.b, By combining the direct transmission (Eq.(5)) and airlight (Eq.(8)) and assuming that the incoming light intensity to a camera is linearly proportional to the camera's pixel values, the scattered light in the atmosphere captured by the camera can be modeled as:

$$\mathbf{I}(\mathbf{x}) = \mathbf{L}\boldsymbol{\rho}(\mathbf{x})e^{-\beta d(\mathbf{x})} + \mathbf{L}_\infty(1 - e^{-\beta d(\mathbf{x})}). \tag{9}$$

The first term is the direct transmission, and the second term is the airlight. The term $\mathbf{I}$ is the image intensity as an RGB color vector [1], while $\mathbf{x}$ is the 2D image spatial location. The term $\mathbf{L}_\infty$ is the atmospheric light that is assumed to be globally constant and independent from location $\mathbf{x}$. The term $\mathbf{L}$ represents the atmospheric light, the camera gain, and the squared distance, $\mathbf{L} = \mathbf{L}_\infty g/d^2$. The term $\boldsymbol{\rho}$ is the reflectance of an object, $\beta$ is the atmospheric attenuation coefficient, and $d$ is the distance between an object and the camera. The term $\beta$ is assumed to independent from wavelengths, which is a common assumption as we are dealing with particles that the size is larger compared with the wavelength of light, such as, fog, haze, aerosol, *etc.* [3]. Moreover, $\beta$ is independent from the spatial image location for homogeneous distribution of atmospheric particles.

In this paper, we denote scene reflection as:

$$\mathbf{R}(\mathbf{x}) = \mathbf{L}\boldsymbol{\rho}(\mathbf{x}), \tag{10}$$

which the estimation of its values is the ultimate goal of dehazing or visibility enhancement, since they indicate the scene that are not affected by medium particles. The term $\mathbf{A}(\mathbf{x})$ represents the airlight, thus

$$\mathbf{A}(\mathbf{x}) = \mathbf{L}_\infty(1 - e^{-\beta d(\mathbf{x})}). \tag{11}$$

Function $t(\mathbf{x})$ represents the transmission, as $t(\mathbf{x}) = e^{-\beta d(\mathbf{x})}$. Hence, the scattering model in Eq.(9) can be written as:

$$\mathbf{I}(\mathbf{x}) = \mathbf{D}(\mathbf{x}) + \mathbf{A}(\mathbf{x}), \tag{12}$$

where $\mathbf{D}(\mathbf{x}) = \mathbf{R}(\mathbf{x})t(\mathbf{x})$, the direct transmission.

The above scattering model assumes the images are three channel RGB images. For gray images, we can write a similar formula by transforming the color vectors to scalar variables:

$$I(\mathbf{x}) = D(\mathbf{x}) + A(\mathbf{x}), \tag{13}$$

where $D(\mathbf{x}) = t(\mathbf{x})R(\mathbf{x})$, with $R(\mathbf{x}) = L\rho(\mathbf{x})$, and, $A(\mathbf{x}) = L_\infty(1 - e^{-\beta d(\mathbf{x})})$.

## III. Survey

In this section, we survey a number of existing researches on vision through atmospheric media. Specifically, methods on depth estimation and/or visibility enhancement. As listed in Table II, we categorize these methods into five categories according to there input and output: (1) depth estimation, (2) multi-images based dehazing, (3) polarizing filters based dehazing, (4) dehazing using known depth, and (5) single image dehazing. For each category, the discussion is done chronologically to show the development of the techniques over time. Since single-image enhancement methods have received the greatest attention recently in the computer vision community, they make up the largest portion of this survey.

### A. Early Work in Depth Estimation

**Cozman-Krotkov 1997** [4] is one of the earliest methods to analyze image of scenes captured in scattering media. The goal in this work is to extract scene depth exploiting the presence of the atmospheric scattering effects. Based on the scattering model in Eq.(9), this approach first obtains the atmospheric light, $L_\infty$, by averaging the sky intensity regions. Then, assuming the reflection of the objects in the scene, $R$, is provided, the transmission can be obtained using:

$$e^{-\beta d(\mathbf{x})} = \frac{I(\mathbf{x}) - L_\infty}{R(\mathbf{x}) - L_\infty}. \tag{14}$$

[1]That is to say we have three sets of equations for wavelength $\lambda$ at red, green and blue channel separately.

TABLE II
AN OVERVIEW OF EXISTING WORKS ON VISION THROUGH ATMOSPHERIC SCATTERING MEDIA.

| **Method** | **Category** | **Known Parameters (Input)** | **Estimating (Output)** | **Key idea** |
|---|---|---|---|---|
| Cozman – Krotkov 1997 | Depth estimation | Single grayscale image I(**x**)<br>Scene reflection R(**x**) | Transmission t(**x**);<br>depth d(**x**); | Direct solving |
| Nayar – Narasimham 1999<br>Method 1 | Depth estimation | Two grayscale images I(**x**) with different scattering coefficients $\beta_1, \beta_2$ | t(**x**), d(**x**) | Comparing different $\beta$ |
| Nayar – Narasimham 1999<br>Method 2 | Depth estimation | Single grayscale image I(**x**)<br>Atmospheric light $L_\infty$ | t(**x**), d(**x**) | Direct solving |
| Nayar – Narasimham 1999<br>Method 3 | Depth estimation | Single RGB image **I**(**x**) | t(**x**), d(**x**) ,<br>Air light: **A**(**x**), | Dichromatic model |
| Nayar – Narasimham 2000 | Multi-images | Two RGB images **I**(**x**)<br>with different weather conditions $\beta_1, \beta_2$ | t(**x**), d(**x**) | Iso – depth: Comparing different $\beta$; colour decomposition |
| Nayar – Narasimham 2003a | Multi-images | Two grayscale or RGB images **I**(**x**) with different weather conditions $\beta_1, \beta_2$ | t(**x**), d(**x**), **A**(**x**) and<br>Scene reflection **R**(**x**) | Iso – depth |
| Caraffa-Tarel 2012 | Multi-images | Stereo Cameras | d(**x**), **R**(**x**) | Depth from scattering; Depth from stereo;<br>Spatial smoothness |
| Li et al. 2015 | Multi-images | Monocular video | t(**x**), d(**x**), **R**(**x**) | Depth from monocular video;<br>Depth from scattering; Photoconsistency |
| Schechner et al. 2001 | Polarizing Filter | Two images with different polarization under same weather condition<br>Image with sky region presented | **A**(**x**), t(**x**), d(**x**), **R**(**x**) | Assuming direct transmission **D**(**x**) has insignificant polarization |
| Schartz et al. 2006 | Polarizing Filter | Two images with different polarization under same weather condition<br>Image with sky region presented | **A**(**x**), t(**x**), d(**x**), **R**(**x**) | Direct transmission **D**(**x**) has insignificant polarization<br>**A**(**x**) and **D** (**x**) are statistically independent |
| Oakley – Satherley 1998 | Known Depth | Single grayscale image I(**x**)<br>Depth d(**x**) | Atmospheric light: $\boldsymbol{L}_\infty$<br>Scattering coefficient: $\beta$<br>R(**x**) | Mean square optimization<br>Colour of the scene is uniform |
| Nayar – Narasimham 2003b<br>Method 1 | Known Depth | Single RGB image **I**(**x**)<br>User specified less hazed and more hazed regions | **R**(**x**) | Dichromatic model |
| Nayar – Narasimham 2003b<br>Method 2 | Known Depth | Single RGB image **I**(**x**)<br>User specified vanishing point, min depth and max depth | **R**(**x**) | Dichromatic model |
| Hautiere et al. 2007 | Known Depth | Single image **I**(**x**)<br>Scene of flat ground | $\boldsymbol{L}_\infty$, **R**(**x**) | Depth from calibrated camera |
| Kopf et al. 2008 | Known Depth | Single image **I**(**x**)<br>Known 3D model | t(**x**), **R**(**x**) | Transmission estimation using averaged texture<br>from same depth |
| Tan 2008 | Single image | Single RGB image **I**(**x**) | $\boldsymbol{L}_\infty$, t(**x**), **R**(**x**) | Brightest value assumption for Atmospheric light $L_\infty$ estimation; Maximal contrast assumption for Scene reflection R(x) estimation |
| Fattal 2008 | Single image | Single RGB image **I**(**x**) | $\boldsymbol{L}_\infty$, t(**x**), **R**(**x**) | Shading and transmission are locally and statistically uncorrelated |
| He et al. 2009 | Single image | Single RGB image **I**(**x**) | $\boldsymbol{L}_\infty$, t(**x**), **R**(**x**) | Dark channel: outdoor objects in clear weather have at least one colour channel that is significantly dark |
| Tarel – Hautiere 2009 | Single image | Single RGB image **I**(**x**) | $\boldsymbol{L}_\infty$, t(**x**), **R**(**x**) | Maximal contrast assumption;<br>Normalized air light is upper-bounded |
| Kratz – Nishino 2009 | Single image | Single RGB image **I**(**x**) | t(**x**), **R**(**x**) | Scene reflection **R**(**x**) and Air light **A**(**x**) are statistically independent; Layer separation |
| Ancuti-Ancuti 2010 | Single image | Single RGB image **I**(**x**) | **A**(**x**), **R**(**x**) | Gray-world colour constancy;<br>Global contrast enhancement |
| Meng et al. 2013 | Single image | Single RGB image **I**(**x**) | $\boldsymbol{L}_\infty$**, t (x), R(x)** | Dark channel for transmission t(**x**) |
| Tang et al. 2014 | Single image | Single RGB image **I**(**x**) | t (**x**), **R**(**x**) | Machine learning of transmission t(**x**) |
| Fattal 2014 | Single image | Single RGB image **I**(**x**) | $\boldsymbol{L}_\infty$**, t (x), R(x)** | Colour line: small image patch has uniform colour and depth but different shading |
| Cai et al. 2016 | Single image | Single RGB image **I**(**x**) | t (**x**), **R**(**x**) | Learning of t(**x**) in CNN framework |
| Berman et al. 2016 | Single image | Single RGB image **I**(**x**) | t (**x**), **R**(**x**) | Non-local haze line; finite colour approximation |

However, the absolute depth, $d(\mathbf{x})$, will still be unknown since the value of $\beta$ is unknown. To resolve this, we need two pixels that have the same value of $\beta$ to obtain the relative depth, such that:

$$\frac{d(\mathbf{x}_i)}{d(\mathbf{x}_j)} = \frac{\log\left(\frac{I(\mathbf{x}_i)-L_\infty}{R(\mathbf{x}_i)-L_\infty}\right)}{\log\left(\frac{I(\mathbf{x}_j)-L_\infty}{R(\mathbf{x}_j)-L_\infty}\right)}, \tag{15}$$

where $\mathbf{x}_i \neq \mathbf{x}_j$. If we have a reference point in the input image whose depth is known, then we can obtain the absolute depth of every pixel in the image. Note that, $R$ is given from an image of exactly the same scene taken in a clear day, though of course, it is a considerably rough approximation; since even in a clear day, an outdoor image is always affected by medium particles, particularly for faraway objects.

**Nayar-Narasimhan 1999** [5] and later [6] propose three different algorithms to estimate depth from hazy scenes. Unlike [4], however, this work does not assume that the reflection, $R$, of the scene without the effects of haze is provided. The first of the three algorithms employs only the direct transmission to estimate the relative depths of light sources from two images taken under different scattering coefficients at nighttime. The idea is to apply the logarithm to the ratio of the pixel intensities of the two input images (where the airlight is assumed to be ignorable at nighttime). Given gray images of nighttime where only the light sources are visible, we compute:

$$K(\mathbf{x}) = \frac{I_1(\mathbf{x})}{I_2(\mathbf{x})} = \frac{D_1(\mathbf{x})}{D_2(\mathbf{x})} = e^{-(\beta_1-\beta_2)d(\mathbf{x})}, \tag{16}$$

where index $1, 2$ indicates the first and second images. The term $D$ is the direct transmission (Eq.(13)). The relative depth from two pairs of pixels can be obtained by:

$$\frac{\log K(\mathbf{x}_i)}{\log K(\mathbf{x}_j)} = \frac{d(\mathbf{x}_j)}{d(\mathbf{x}_j)}. \tag{17}$$

This relative depth is not for the entire image, but only for the light source regions.

The second algorithm is to estimate the absolute depth from a single airlight image. It assumes that the atmospheric light, $L_\infty$, and achromatic airlight, $A(\mathbf{x})$, are given:

$$\log\left(\frac{L_\infty - A(\mathbf{x})}{L_\infty}\right) = -\beta d(\mathbf{x}). \tag{18}$$

The problem with this algorithm is with regards obtaining the airlight, which is discussed in the third algorithm.

The third algorithm treats the problem of depth estimation as a color decomposition of the scattering model (Eq.(9)). The method decomposes the input images into the chromaticity of the direct attenuation and the chromaticity of the airlight; the latter is identical to the chromaticity of the atmospheric light. Chromaticity is generally defined as a normalized color, where $R/(R+G+B)$, and $G/(R+G+B)$. This is a unit color vector and can be used to convert the model of Eq.(9) into a chromaticity based formulation:

$$\mathbf{I}(\mathbf{x}) = p(\mathbf{x})\hat{\mathbf{D}}(\mathbf{x}) + q(\mathbf{x})\hat{\mathbf{A}}(\mathbf{x}), \tag{19}$$

where $\hat{\mathbf{D}}$ and $\hat{\mathbf{A}}$ are the chromaticity values of the direct transmission and the airlight. The terms $p$ and $q$ are the magnitude of the direct transmission and the airlight, respectively. The paper calls the equation the *dichromatic scattering model*, where the word dichromatic is borrowed from [24] due to the similarity of the models.

In the RGB space, the two chromaticity vectors (the direct transmission chromaticity and airlight chromaticity) will create a plane. The same idea had been discussed in [24][25] for the dichromatic model of specular highlights. Accordingly, given a pixel and the two chromaticity values, we can immediately calculate the magnitude of the airlight, which consequently gives us the absolute depth map by employing Eq.(18). In this algorithm, the chromaticity of the direct transmission is assumed to be given from a clear day image of exactly the same scene, and the airlight chromaticity is computed from a known atmospheric light.

### *B. Multiple Images*

**Narasimhan-Nayar 2000** [7] extends the analysis of the dichromatic scattering model of [5] in Eq.(19) by using multiple images of the same scene taken in different haze density. The method works by supposing there are two images taken from the same scene, which share the same color of atmospheric light, but different colors of direct transmission. From this, two planes can be formed in the RGB space that intersect to each other. In their work [7] utilizes the intersection to estimate the atmospheric light chromaticity, $\hat{\mathbf{A}}$, which is similar to Tominaga and Wandell's method [26] for estimating a light color from specular reflection. The assumption that the images of the same scene have different colors of direct transmission, however, might produce inaccurate estimation since, in many cases, the colors of the direct transmission of the same scene are similar.

The method then introduces the concept of iso-depth, which is the ratio of the direct transmission magnitudes under two different weather conditions. Referring to Eq.(19), and applying it to two images, we have:

$$\frac{p_2(\mathbf{x})}{p_1(\mathbf{x})} = \frac{L_{\infty 2}}{L_{\infty 1}} e^{-(\beta_2-\beta_1)d(\mathbf{x})}, \tag{20}$$

where $p$ is the magnitude of the direct transmission. From this equation, we can infer that if two pairs of pixels have the same ratio, then they must have the same depth: $\frac{p_2(\mathbf{x}_i)}{p_1(\mathbf{x}_i)} = \frac{p_2(\mathbf{x}_j)}{p_1(\mathbf{x}_j)}$. To calculate these ratios, the method provides a solution by utilizing the analysis of the planes formed in the RGB space by the scattering dichromatic model in Eq.(19).

Having obtained the ratios for all pixels, the method proceeds with the estimation of the scene structure, which is calculated by:

$$(\beta_2 - \beta_1)d(\mathbf{x}) = \log\left(\frac{L_{\infty 2}}{L_{\infty 1}}\right) - \log\left(\frac{p_2(\mathbf{x})}{p_1(\mathbf{x})}\right). \tag{21}$$

To be able to estimate the depth, the last equation requires the knowledge of the values of $L_{\infty 1}$ and $L_{\infty 2}$, which are obtained by solving the following linear equation:

$$c(\mathbf{x}) = L_{\infty 2} - \frac{p_2(\mathbf{x})}{p_1(\mathbf{x})} L_{\infty 1}, \tag{22}$$

where $c$ is the magnitude of a vector indicating the distance between the origin of vector $\mathbf{I}_1$ to the origin of vector $\mathbf{I}_2$ in the direction of the airlight chromaticity in RGB space. While, $\frac{p_2(\mathbf{x})}{p_1(\mathbf{x})}$ is the ratio, which had been computed.

For the true scene color restoration, employing the estimated atmospheric light, the method computes the airlight magnitude of Eq.(19) using:

$$q(\mathbf{x}_i) = L_\infty \left(1 - e^{-\beta d(\mathbf{x}_i)}\right), \tag{23}$$

where:

$$\beta d(\mathbf{x}_i) = \beta d(\mathbf{x}_j) \left(\frac{d(\mathbf{x}_i)}{d(\mathbf{x}_j)}\right), \tag{24}$$

and $\frac{d(\mathbf{x}_i)}{d(\mathbf{x}_j)}$ is computable using Eq.(21). $\beta d(\mathbf{x}_j)$ is a chosen reference point. This is obtained by assuming there is at least a pixel that the true value of the direct transmission, $\mathbf{D}$, is known (*e.g.* a black object); since, in this case $\mathbf{I}(\mathbf{x}) = \mathbf{A}(\mathbf{x})$, and $\beta d(\mathbf{x})$ can be directly computed. The method also proposes how to find such a pixel automatically. Note that, knowing the value of $q(\mathbf{x}_i)$ in Eq.(23) enables us to dehaze the images straightforward manner.

**Narasimhan-Nayar 2003** In a subsequent publication, Narasimhan and Nayar [8] introduce a technique that work for gray or colored images: contrast restoration of iso-depth regions, atmospheric light estimation, and contrast restoration.

In the contrast restoration of iso-depth regions, the method forms an equation that assumes the depth segmentation is provided (*e.g.* manually by the user) and the atmospheric light is known:

$$\rho(\mathbf{x}_i) = 1 - \left(\sum_j 1 - \sum_j \rho(\mathbf{x}_j)\right) \frac{L_\infty - I(\mathbf{x}_i)}{\sum_j (L_\infty - I(\mathbf{x}_j))}, \tag{25}$$

where the sums are over the same depth regions. As can be seen in the equation, $\rho(\mathbf{x}_i)$ can be estimated up to a linear factor $\sum_j \rho(\mathbf{x}_j)$. By setting $\rho^{min} = 0$ and $\rho^{max} = 1$ and adjusting the value $\sum_j \rho(\mathbf{x}_j)$, the contrast of regions with the same depth can be restored.

To estimate the atmospheric lights, the method utilizes two gray images of the same scene yet different atmospheric lights. Based on the scattering model in Eq.(9), scene reflectance, $\rho$, is eliminated. The two equations representing the two images can be transformed into:

$$I_2(\mathbf{x}) = \left[\frac{L_{\infty 2}}{L_{\infty 1}} e^{-(\beta_2 - \beta_1) d(\mathbf{x})}\right] I_1(\mathbf{x}) + \left[L_{\infty 2}\left(1 - e^{-(\beta_2 - \beta_1) d(\mathbf{x})}\right)\right], \tag{26}$$

where index $1, 2$ indicate image 1 and 2, respectively. From the equation, a two dimensional space can be formed, where $I_1$ is the $x$-axis, and $I_2$ is the $y$-axis. For pixels representing objects that have the same depth, $d$, yet different reflectance, $\rho$, will form a line. As a result, if we have different depths, then there will be a few different lines, which intersect at $(L_{\infty 1}, L_{\infty 2})$. The lines that represent pixels with the same depth can be detected using the Hough transform. As for the contrast restoration or dehazing, the method is the same as that proposed in Narasimhan and Nayar's multiple images method [7].

**Caraffa-Tarel 2013** [27] introduces a dehazing method that includes depth estimation based on stereo cameras. The motivation is that both airlight and disparity from stereo can indicate the scene depths. The goal of the method is to estimate the depth and restored visibility. To achieve the goal, the objective function for the data term is composed of two main factors, the stereo and fog data terms:

$$E_{data} = \sum_{\mathbf{x}} \alpha E^{data}_{stereo}(\mathbf{x}) + (1-\alpha) E^{data}_{fog}(\mathbf{x}), \tag{27}$$

where $\alpha$ is the weighting factor, and

$$E^{data}_{stereo}(\mathbf{x}) = f_s\left(\frac{\mathbf{I}_L(x,y) - \mathbf{I}_R(x - \delta(x,y), y)}{\sigma_s}\right) \tag{28}$$

$$\begin{aligned} E^{data}_{fog}(\mathbf{x}) = {} & |\mathbf{I}_L(x,y) - \mathbf{R}_L(x,y) e^{-\beta \frac{b}{\delta(x,y)}} \\ & + \mathbf{L}_\infty (1 - e^{-\beta \frac{b}{\delta(x,y)}})|^2 \\ & |\mathbf{I}_R(x,y) - \mathbf{R}_R(x,y) e^{-\beta \frac{b}{\delta(x,y)}} \\ & + \mathbf{L}_\infty (1 - e^{-\beta \frac{b}{\delta(x,y)}})|^2, \end{aligned} \tag{29}$$

where indexes $L, R$ indicate the left and right images, respectively. Function $f_s$ relates to the distribution of noise with variance $\sigma_s$. The term $\delta$ is the stereo disparity, and $b$ relates to camera parameters such as baseline and focal length.

Aside from the data terms, the method utilizes prior terms, which is basically the spatial smoothness term for the estimated disparity, $\delta$, and the estimated $\mathbf{R}_L$. The optimization is done by decoupling estimation of the stereo and fog terms. Specifically, it first minimizes the fog term by holding $\delta$ fixed (where $\delta$ is initialized in the first iteration). Then, having $\mathbf{R}_L$ and $\mathbf{R}_R$, it minimizes the stereo term to obtain $\delta$. This is done iteratively until convergence. While this paradigm is reasonable, this work does not address whether the minimization of the fog term can produce better $\mathbf{R}$, and in turn whether the minimization of the stereo can produce better $\delta$. As such, it is not entirely clear whether the decoupling process can support each other.

**Li *et al.* 2015** [9] jointly estimates scene depth and enhance visibility in a foggy video, which unlike Caraffa-Tarel's method [27] uses a monocular video. Following the work of Zhang *et al.* [28], it estimates the camera parameters and the initial depth of the scene, which is erroneous particularly for dense fog regions due to the photoconsistency problem in the data term. In clear scenes, photoconsistency can be achieved by measuring the RGB distance between a pixel in one frame and its estimated corresponding pixel in another frame, however it will be inaccurate when the region is affected by dense fog. Indeed, any existing defogging methods can be used to help improve the intensity values, yet the paper claims existing methods are intended to handle a single image, and when applied to a video sequence, the results will be inconsistent from frame to frame, causing the photoconsistency term to be unstable. To resolve the problem, Li *et al.*'s method [9] introduces a new photoconsistency term:

$$E_p(d_n) = \frac{1}{|\mathcal{N}(n)|} \sum_{n' \in \mathcal{N}(n)} \sum_{\mathbf{x}} \|\hat{I}_{n'}(\mathbf{x}) - I_{n'}(l_{n \to n'}(\mathbf{x}, d_n(\mathbf{x})))\|, \tag{30}$$

where $l_{n \to t'}(\mathbf{x}, d_n(\mathbf{x})$ projects the pixel x with inverse depth $d_n(\mathbf{x})$ in frame $n$ to frame $n'$. The intensity, $\hat{I}_{n'}(\mathbf{x}) =$

$(I_n(\mathbf{x})-L_\infty)\frac{\pi_{n\to n'}(\mathbf{x},t_n(\mathbf{x}))}{t_n(\mathbf{x})}+L_\infty$, is a synthetic intensity value obtained from the transmission, $t_n$, which is computable by knowing $d_n$ (note that, in the paper, the scattering coefficient $\beta$ and the atmospheric light, $L_\infty$, are estimated separately). The projection function $\pi_{n\to n'}(\mathbf{x},t_n(\mathbf{x}))$ computes the corresponding transmission in the $n'$-th frame for the pixel $\mathbf{x}$ in the $n$-th frame with transmission $t_n(\mathbf{x})$. The denominator $\mathcal{N}(t)$ represents the neighboring frames of frame $n$ and $|\mathcal{N}(n)|$ is the number of neighboring frames. By having $\beta(\mathbf{x})$ estimated separately, $t_n(\mathbf{x})$ depends only on $d_n(\mathbf{x})$, and thus $d_n$ is the only unknown in the last equation. The whole idea in the photoconsistency term here is to generate a synthetic intensity value of each pixel from known depth, $d$ , atmospheric light, $L_\infty$, and the particle scattering coefficient, $\beta$. Note that, the paper assumes $\beta$ and $L_\infty$ are uniform across the video sequence. Therefore, if those three values are correctly estimated, the generated synthetic intensity values must be correct.

Aside from the photoconsistency term, the method also uses Laplacian smoothing as the transmission smoothness prior. The whole framework is an integrated framework, where after a few iterations, the outcomes are estimated depth maps and defogged images.

### C. Polarizing Filter

**Schechner** *et al.* **2001** addresses the issue appeared in the work of Narasimhan and Nayar [7], where it requires at least two images of the same scene taken under different particle densities (*i.e.* we have to wait until the fog density changes considerably). Unlike [7], Schecher *et al.*'s [10] uses multiple images captured using polarizing filters, which does not require the fog density to change.

The main assumption employed in this polarized-based method is that the direct transmission has insignificant polarization, and thus the polarization of the airlight dominates the observed light. Based on this, the maximum intensity occurs when airlight passes the through the filter. This can be obtained when:

$$I^{\max}(\mathbf{x}) = D(\mathbf{x})/2 + A^{\max}(\mathbf{x}), \tag{31}$$

where $D$ and $A$ are the direct transmission and the airlight, respectively. The minimum intensity (*i.e.* when the filter can block the airlight at its best) is when:

$$I^{\min}(\mathbf{x}) = D(\mathbf{x})/2 + A^{\min}(\mathbf{x}). \tag{32}$$

Adding up the two states of the polarization, we obtain: $I(\mathbf{x}) = I^{\max}(\mathbf{x}) + I^{\min}(\mathbf{x})$. Based on this, the method estimates the atmospheric light from a sky region and computes its degree of polarization:

$$P = \frac{L_\infty^{\max} - L_\infty^{\min}}{L_\infty^{\min} + L_\infty^{\max}}, \tag{33}$$

and then, estimate the airlight for every pixel:

$$A(\mathbf{x}) = \frac{I^{\max}(\mathbf{x}) - I^{\min}(\mathbf{x})}{P}. \tag{34}$$

Based on the airlight, the method computes the transmission: $e^{-\beta d(\mathbf{x})} = 1 - \frac{A(\mathbf{x})}{L_\infty}$, and finally obtains the dehazing result $R(\mathbf{x}) = [I(\mathbf{x}) - A(\mathbf{x})]\, e^{\beta d(\mathbf{x})}$. To obtain the maximum and the minimum intensity values, the filter needs to be rotated either automatically or manually.

**Shwartz** *et al.* **2006** [11] uses the same setup proposed by Schechner *et al.*'s [10] but removes the assumption that sky regions are present in the input image. Instead, this method estimates the color of the airlight and of the direct transmission by applying independent component analysis (ICA):

$$\begin{bmatrix} A \\ D \end{bmatrix} = \mathbb{W}\begin{bmatrix} I^{max} \\ I^{min} \end{bmatrix} \tag{35}$$

$$\mathbb{W} = \begin{bmatrix} 1/P & -1/P \\ (P-1)/P & (P+1)/P \end{bmatrix}. \tag{36}$$

In this case, the challenge lies in estimating $\mathbb{W}$ given $[I^{max}, I^{min}]^T$ to produce $D$ and $A$ accurately.

The method claims that while the airlight and direct transmission are in fact statistically not independent and certain transformations such as a wavelet transformation can relax the dependence. The method therefore transforms the input data using a wavelet transformation, solves the ICA problem by using an optimization method in the wavelet domain. Aside from $P$, the method also needs to estimate $L_\infty$, which is done by labeling certain regions manually to have two pixels that have the same values of the direct transmission yet different values of the airlight.

### D. Known Depth

**Oakley-Satherley 1998** [12] is one of the early methods dealing with visibility enhancement in a single foggy image. The enhancement is done in two stages: parameter estimation followed by contrast enhancement. The basic idea of the parameter estimation is to employ the sum of squares method to minimize an error function, between the image intensity and some parameters of the physical model, by assuming the reflectance of the scene can be approximated by a single value representing the mean of the scene reflectance. With these assumptions, the minimization is done to estimate three global parameters: the atmospheric light ($L_\infty$), the mean reflectance of the whole scene $\bar{\rho}$, and the scattering coefficient, $\beta$:

$$Err = \sum_{\mathbf{x}}^{M} \left( I(\mathbf{x}) - L_\infty \left( 1 + (\bar{\rho} - 1) e^{-\beta d(\mathbf{x})} \right) \right)^2. \tag{37}$$

The last equation assumes that $L = L_\infty$. Having estimated the three global parameters by minimizing function $Err$, the airlight is then computed using:

$$A(\mathbf{x}) = L_\infty(1 - e^{-\beta d(\mathbf{x})}). \tag{38}$$

Consequently, the end result is obtained by computing:

$$R(\mathbf{x}) = \left( L_{max} \left( \frac{I(\mathbf{x}) - A(\mathbf{x})}{L_\infty} e^{\beta d(\mathbf{x})} \right) \right)^{\frac{1}{2.2}}, \tag{39}$$

where $L_{max}$ is a constant depending on the maximum gray level of the image display device, and $\frac{1}{2.2}$ is to compensate the gamma correction.

The main drawbacks of this method are the assumption that the depth of the scene is known, and the mean reflectance for the whole image is used in the minimization and in computing the airlight. The latter is acceptable if the color of the scene is somehow uniform, which is not the case for general scenes. Tan and Oakley's [29] extended the work of Oakley and Satherley [12] to handle color images by taking into account a colored scattering coefficient and colored atmospheric light.

**Narasimhan-Nayar 2003** [30] proposes several methods based on a single input image; however due to the ill-posed nature of the problem, the methods requires some user interaction. The first method requires the user to select a region with less haze and a region with more haze of the same reflection as the first one's. From these the two inputs, the approach estimates the dichromatic plane and dehaze pixels that have the same color as the region with less haze. This method assumes the pixels represent scene points that have the same reflection. The second method asks the user to indicate the vanishing point and to input the maximum and minimum distance from the camera. This information is used to interpolate the distance to estimate the clear scene in between. The interpolation is a rough approximation, since depth can be layered and not continuous. To resolve layered scenes, the third method is introduced, which requires depth segmentation that can be done through satellite orthographic photos of buildings.

**Hautiere** ***et al.*** **2007** [13] proposes a method to dehaze a scene from a single image that assumes a flat world (*i.e.* only flat ground without trees or other objects) and known camera properties including its height from the ground. These assumptions are necessary to estimate the attenuation factor and the depth, which are expressed as:

$$d = \frac{a}{y - y_h}, \qquad \text{if } v > v_h, \tag{40}$$

where $a = \frac{H\alpha}{\cos^2\theta}$. The term $H$ is the height of the camera, $y$ is the $y$-axis of the image coordinates, $\theta$ is the angle between the optical axis of the camera and the horizon line. $y_h$ is the horizon line. The term $\alpha = f/w$, with $f$ is the focal length and $w$ is the height of a pixel. Hence, $d$ in the scattering model (Eq.(9)) is replaced by $\frac{a}{y-y_h}$. By taking the derivative of the model with regard to $y$, we obtain:

$$\frac{d^2 I}{dy^2} = \beta\psi(y) e^{-\beta\frac{a}{y-y_h}}\left(\frac{\beta a}{y-y_h} - 2\right), \tag{41}$$

where $\psi(y) = \frac{a(R-L_\infty)}{(y-y_h)^3}$. Setting $\frac{d^2 I}{dy^2} = 0$ produces:

$$\beta = \frac{2(y_i - y_h)}{a}. \tag{42}$$

If we can find the value of $y_i$, which is the inflection point in the vertical axis of the image, then we can obtain the value of $\beta$ since the method assumes we can know the horizon line, $y_h$, from the input image. By knowing $y_i$ and $\beta$ we obtain:

$$L_\infty = I_i + \frac{y_i - y_h}{2}\frac{dI}{dy}_{|y=y_i}, \tag{43}$$

$$R = I_i - \left(1 - e^{-\beta d_i}\right)\left(\frac{y_i - y_h}{2e^{-\beta d_i}}\right)\frac{dI}{dy}_{|y=y_i}, \tag{44}$$

where $R$ indicates the reflection of the ground.

To find the inflection point location, $y_i$, the method utilizes the median intensity of each line of a vertical band, which should be only located at the homogeneous area and the sky.

To relax the flat world assumption, which does not apply to trees, vehicles, houses, or any objects in the scene, the method employs depth heuristics, such as a cylindrical scene. Due to all these constraints, the method works in its full potential for scenes dominated by flat planes (*e.g.* rural road scenes).

**Kopf** ***et al.*** **2008** [14] attempts to overcome the dehazing problem by utilizing the information provided by an exact 3D model of the input scene and the corresponding model textures (obtained from Landsat data). The main task is to estimate the transmission, $\exp(-\beta d(\mathbf{x}))$, and the atmospheric light, $\mathbf{L}_\infty$.

Since, it has the 3D model of the scene, it can collect the average model texture intensity of certain depths ($\hat{I}_h(\mathbf{x})$) from the Landsat data and the corresponding average haze intensity ($\hat{I}_m(\mathbf{x})$) of the same depths from the input image. The two average intensity values can be used to estimate the transmission assuming $\mathbf{L}_\infty$ is known:

$$t(\mathbf{x}) = \frac{\hat{I}_h - L_\infty}{C\hat{I}_m - L_\infty}, \tag{45}$$

where $C$ is a global correction vector. By comparing this equation with Eq.(14), we see that $C\hat{I}_m$ attempts to substitute $R$, the scene reflectance without the influence of haze. In this method, $C$ is computed from:

$$C = \frac{F_h}{\text{lum}(F_h)} / \frac{F_m}{\text{lum}(F_m)}, \tag{46}$$

where $F_h$ is the average of $I_h(\mathbf{x})$ with $z < z_F$ with $z_F$ is set to 1600 meters, and $F_m$ is the average of the model texture. The function lum($c$) is the luminance of a color $c$.

The method suggests that $\mathbf{L}_\infty$ is estimated by collecting the average background intensity for pixels whose depth is more than a certain distance ($> 5000m$) from both the input image and the model texture image.

### *E. Single-Image Methods*

**Tan 2008** . A milestone in single-image visibility enhancement was made with the publications of Tan [15] and Fattal [16] that can automatically dehaze a single image without additional information such as known geometrical information or user input. Given an input image, Tan's method [15] estimates the atmospheric light, $\mathbf{L}_\infty$ from the brightest pixels in the input image, and normalize the color of the input image, from $\mathbf{I}$ to $\tilde{\mathbf{I}}$ by dividing $\mathbf{I}$ by the chromaticity of $\mathbf{L}_\infty$, element-wise. The chromaticity of $\mathbf{L}_\infty$ is the same as $\hat{\mathbf{A}}$ in Eq.(19). By doing this, the airlight $\mathbf{A}$, can be transformed from color vectors into scalars, $A$. Hence, the visibility enhancement problem can be solved if we know the scalar value of the airlight, $A$, for every pixel:

$$e^{\beta d(\mathbf{x})} = \frac{\sum_c \mathbf{L}_{2c}}{A(\mathbf{x})\sum_c \mathbf{L}_{2c}}, \tag{47}$$

$$\tilde{\mathbf{R}}(\mathbf{x}) = \left(\tilde{\mathbf{I}}(\mathbf{x}) - A(\mathbf{x})\begin{bmatrix}1\\1\\1\end{bmatrix}\right)e^{\beta d(\mathbf{x})}, \tag{48}$$

where $c$ represents the index of RGB channels, and $\tilde{\mathbf{R}}$ is the light normalized color of the scene reflection, $\mathbf{R}$. The values of $A$ range from 0 to $\sum_c \mathbf{L}_{2c}$. The key idea of the method is to find a value of $A(\mathbf{x})$ from that range that maximizes the local contrast of $\tilde{\mathbf{R}}(x)$. The local contrast is defined as:

$$\text{Contrast}(\tilde{\mathbf{R}}(\mathbf{x})) = \sum_{x,c}^{S} |\nabla \tilde{\mathbf{R}}_c(\mathbf{x})|, \tag{49}$$

where $S$ is a local window whose size is empirically set to $5 \times 5$. It was found that the correlation between the airlight and the contrast is convex.

The problem can be casted into a Markov Random Field (MRF) framework and optimized using graphcuts to estimate the values of the airlight across the input image. The method works for both color and gray images and was shown able to handle relatively thick fog. One of the drawbacks of the method is the appearance of halos around depth discontinuity due to the local-window based operation. Another drawback is that when the input regions have no textures, the quantity of local contrast will be constant even when the airlight value changes. Prior to the 2008 publication, Tan *et al.*[31] had introduced a fast single dehazing method that uses a color constancy method [32] to estimate the color of the atmospheric light, and utilizes the $Y$ channel of the $YIQ$ color space as an approximation to do the dehazing.

**Fattal 2008** [16] is based on the idea that the shading and transmission functions are locally and statistically uncorrelated. From this, the work derives the shading and transmission functions from Eq.(9):

$$l^{-1}(\mathbf{x}) = \frac{1 - I_A(\mathbf{x})/||\mathbf{L}_\infty||}{} + \frac{\eta}{||\mathbf{L}_\infty||}, \tag{50}$$

$$t(\mathbf{x}) = 1 - I_A(\mathbf{x}) - \frac{\eta I_{R'}(\mathbf{x})}{||\mathbf{L}_\infty||}, \tag{51}$$

where $l(\mathbf{x})$ is the shading function and $t(\mathbf{x})$ is the transmission function. The definitions of $I_A$ and $I_{R'}$ is as follows:

$$I_A(\mathbf{x}) = \frac{\langle \mathbf{I}(\mathbf{x}), \mathbf{L}_\infty \rangle}{||\mathbf{L}_\infty||}, \tag{52}$$

$$I_{R'}(\mathbf{x}) = \sqrt{||\mathbf{I}_x||^2 - I_A^2(\mathbf{x})}. \tag{53}$$

Assuming $\mathbf{L}_\infty$ can be obtained from the sky regions, $\eta$ is estimated by assuming the shading and the transmission functions are statistically uncorrelated over a certain region $\Omega$. This implies that $C_\Omega(l^{-1}, t) = 0$, where function $C_\Omega$ is the sample covariance. Hence, $\eta$ can be defined based on $C_\Omega(l^{-1}, t) = 0$:

$$\eta(\mathbf{x}) = \frac{C_\Omega\left(I_A(\mathbf{x}), h(\mathbf{x})\right)}{C_\Omega\left(I_{R'}(\mathbf{x}), h(\mathbf{x})\right)}, \tag{54}$$

where $h(\mathbf{x}) = (||\mathbf{L}_\infty|| - I_A(\mathbf{x}))/I_{R'}(\mathbf{x})$. Obtaining the values of $t(\mathbf{x})$ and $\mathbf{L}_\infty$ will eventually solve the estimation of the scene reflection, $\mathbf{R}(\mathbf{x})$.

The success of the method relies on whether the statistical decomposition of shading and transmission can be optimum, and whether they are truly independent. Moreover, while it works for haze, the approach was not tried on foggy scenes.

**He** *et al.* **2009**. The work in [17], [33] observed an interesting phenomenon of outdoor natural scenes with clear visibility. They found that most outdoor objects in clear weather have at least one color channel that is significantly dark. They argue that this is because natural outdoor images are colorful (*i.e.* the brightness varies significantly in different color channels) and full of shadows. Hence, they define a dark channel as:

$$R^{dark} = \min_{y \in \Omega(\mathbf{x})} \left( \min_{c \in \{R,G,B\}} R^c(y) \right). \tag{55}$$

Because of the observation that, $R^{dark} \rightarrow 0$, He *et al.* c̃itehe2009 refer to this as the *dark channel prior*.

The dark channel prior is used to estimate the transmission as follows. Based on Eq.(9), we can express:

$$\frac{I^c(\mathbf{x})}{L^c_\infty} = t(\mathbf{x}) \frac{R^c(\mathbf{x})}{L^c_\infty} + 1 - t(\mathbf{x}). \tag{56}$$

Assuming that we work on a local patch, $\Omega(\mathbf{x})$ and $t(\mathbf{x})$ are constant within the patch, $\tilde{t}(\mathbf{x})$, then we can be written as:

$$\min_{y \in \Omega(\mathbf{x})} \left( \min_c \frac{I^c(\mathbf{x})}{L^c_\infty} \right) = \tilde{t}(\mathbf{x}) \min_{y \in \Omega(\mathbf{x})} \left( \min_c \frac{R^c(\mathbf{x})}{L^c_\infty} \right) + 1 - \tilde{t}(\mathbf{x}), \tag{57}$$

and consequently, due to the dark channel prior:

$$\tilde{t}(\mathbf{x}) = 1 - \min_{y \in \Omega(\mathbf{x})} \left( \min_c \frac{I^c(\mathbf{x})}{L^c_\infty} \right), \tag{58}$$

where $\mathbf{L}_\infty$ is obtained by picking the top 0.1 % brightest pixels in the dark channel. Finally, to have a smooth and robust estimation of $t(\mathbf{x})$ that can avoid the halo effects due to the use of patches, the method employs the closed-form solution of matting [34].

**Tarel-Hautiere 2009** . One of the drawbacks of the previous methods [15] [16] [17] [33] is the computation time. The methods cannot be applied for real time applications, where the depths of the input scenes change from frame to frame. Tarel and Hautiere [35] introduce a fast visibility restoration method whose complexity is linear to the number of image pixels. Inspired by the contrast enhancement [15], they observed that the value of the normalized airlight, $A(\mathbf{x})$ (where the illumination color is now pure white), is always less than $W(\mathbf{x})$, where $W(\mathbf{x}) = \min_c(\tilde{I}^c(\mathbf{x}))$. Note that, $\tilde{I}^c$ is the pixel intensity value of color channel $c$ after the light normalization. Since it takes time to find the optimum value of $A(\mathbf{x})$, the idea of estimating $A(\mathbf{x})$ rapidly is based on a heuristic method:

$$M(\mathbf{x}) = \text{median}_{\Omega(\mathbf{x})}(W)(\mathbf{x}), \tag{59}$$

$$S(\mathbf{x}) = M(\mathbf{x}) - \text{median}_{\Omega(\mathbf{x})}(|W - M|)(\mathbf{x}), \tag{60}$$

$$A(\mathbf{x}) = \max\left(\min(pS(\mathbf{x}), W(\mathbf{x}), 0\right), \tag{61}$$

where $\Omega(\mathbf{x})$ is a patch centered at $x$, and $p$ is a constant value, chosen empirically. The last equation means $0 \leq A(\mathbf{x}) \leq W(\mathbf{x})$. The method utilizes a bilateral filter or median filter to help produce a smooth estimation of the airlight $A(\mathbf{x})$.

**Kratz-Nishino 2009** [36] and later [37] offer a new perspective on the dehazing problem. This work poses the problem in the framework of a factorial MRF [38], which consists of a single observation field (the input hazy image), and two separated hidden fields (the albedo and the depth fields). Thus, the idea of the method is to estimate the depth and albedo by assuming that the two are statistically independent. First, it transforms the model in Eq.(9) to:

$$\log\left(1 - \frac{I^c(\mathbf{x})}{L^c_\infty}\right) = \log(1 - \rho^c(\mathbf{x})) - \tilde{d}(\mathbf{x}), \quad (62)$$

$$\tilde{I}^c(\mathbf{x}) = C^c(\mathbf{x}) + D(\mathbf{x}), \quad (63)$$

where $c$ is the index of the color channel, $C^c(\mathbf{x}) = \log(1 - \rho^c(\mathbf{x}))$, and $D(\mathbf{x}) = -\tilde{d}(\mathbf{x})$, and $\tilde{d}(\mathbf{x}) = \beta d(\mathbf{x})$. Hence, in terms of the factorial MRF, $\tilde{I}^c$ is the observed field, and $C^c$ and $D$ are the two separated hidden fields. Each node in the MRF will connect to the corresponding node in the observed field and to its neighboring nodes within the same field. The goal is then to estimate the value of $C^c$ for all color channels and the depth, $D$. The objective function consists of the likelihood and the priors $C^c$ and $D$. The prior of $C^c$ is based on the exponential power distribution of the chromaticity gradients (from natural images); while the prior of $D$ is manually selected from a few different models, depending on the input scene (*e.g.* either cityscape, terrain, *etc.*). To solve the decomposition problem, the method utilizes an EM algorithm that decouples the estimation of the two hidden fields. In each step, graphcuts is used to optimize the values, resulting in a high computational cost. To make the iteration more efficient a good initializations are required. The initialization for the depth is done by computing:

$$D_{init}(\mathbf{x}) = \max_{c\in R,G,B}(\tilde{I}^c(\mathbf{x})), \quad (64)$$

which means that the depth are initialized from the brightest in the color channel, which is the upper bound of the depth. This last equation is in essence the same as the dark channel prior [17] [33], and the paper [37] offers a different interpretation of the dark channel prior from the viewpoint of depth, namely, the dark channel prior works because it is computed from the upper-bound of the depth and not because nature has many shadows or varying colors, which is the reasoning of He *et al.* [33].

**Ancuti-Ancuti 2010**. The works [39] [40] propose a method based on image fusion. First, the method splits the input image into two components: a white-balanced image, $\mathbf{I}_1$, by using the gray-world color constancy method [41], and a global contrast enhanced image, $\mathbf{I}_2$ , which is calculated by $\mathbf{I}_2(\mathbf{x}) = \gamma(\mathbf{I}(\mathbf{x}) - \bar{\mathbf{I}})$, where $\bar{\mathbf{I}}$ is the average intensity of the whole input image and $\gamma$ is a weighting factor. From both $\mathbf{I}_1$ and $\mathbf{I}_2$, the weights in terms of the luminance, chromaticity, and saliency are calculated. Based on the weights, the output of the dehazing algorithm is

$$\tilde{w^1}(\mathbf{x})\mathbf{I}_1 + \tilde{w^2}(\mathbf{x})\mathbf{I}_2, \quad (65)$$

where $\tilde{w}^k$ is the normalized weights with index $k$ is either 1 or 2, such that $w^k(\mathbf{x}) = w^k_l w^k_c w^k_s$ and $\tilde{w}^k = w^k/\sum_{k=1}^2 w^k$. The subscripts $l, c, s$ represent luminance, chromaticity and saliency, respectively. The weights' definitions are as follows:

$$w^k_l(\mathbf{x}) = \sqrt{\frac{1}{3}\sum_{c\in R,G,B}\left(I^k_c(\mathbf{x}) - L^k(\mathbf{x})\right)^2}, \quad (66)$$

$$w^k_c(\mathbf{x}) = \exp\left(-\frac{\left(S^k(\mathbf{x}) - S^k_{max}\right)^2}{2\sigma^2}\right), \quad (67)$$

$$w^k_s(\mathbf{x}) = ||I^k_\omega(\mathbf{x}) - I^k_\mu||, \quad (68)$$

where $L^k(\mathbf{x})$ is the average of the intensity in the three color channels. The term $S$ is the saturation value (*e.g.* the saturation in the HSI color space). The term $\sigma$ is set 0.3 as default. The term $S_{max}$ is a constant, where for the HSI color space, it would be 1. The term $I^k_\mu$ is the arithmetic mean pixel value of the input, and $I^k_\omega$ is the blurred input image. The method produces good results, however the reasoning behind using the two images ($\mathbf{I}_1$ and $\mathbf{I}_2$) and the three weights are not fully explained and need further investigation. The fusion approach was also applied to underwater vision [42].

**Meng** ***et al.*** **2013** [23] extends the idea of the dark channel prior in determining the initial values of transmission, $t(\mathbf{x})$, by introducing its lower bound. According to Eq.(9), $t(\mathbf{x}) = (A^c - I^c(\mathbf{x}))/(A^c - R^c(\mathbf{x}))$. As a result, the lower bound of the transmission, denoted as $t_b(\mathbf{x})$ can be defined as:

$$t_b(\mathbf{x}) = \frac{A^c - I^c(\mathbf{x})}{A^c - C^c_0}, \quad (69)$$

where $C^c_0$ is a small scalar value. Since $C^c_0$ is smaller or equal than $R^c(\mathbf{x})$, then $t_b(\mathbf{x}) \leq t(\mathbf{x})$. To anticipate a wrong estimation of $\mathbf{A}$, such as when the value of $A^c$ is smaller than $I^c$, the second definition of $t_b(\mathbf{x})$ is expressed as:

$$t_b(\mathbf{x}) = \frac{A^c - I^c(\mathbf{x})}{A^c - C^c_1}, \quad (70)$$

where $C^c_1$ is a scalar value, larger than the possible values of $A^c$ and $I^c$. Combining the two definitions, we obtain:

$$t_b(\mathbf{x}) = \min\left(\max_{c\in R,G,B}\left(\frac{A^c - I^c(\mathbf{x})}{A^c - C^c_0}, \frac{A^c - I^c(\mathbf{x})}{A^c - C^c_1}\right), 1\right). \quad (71)$$

Assuming the transmission is constant for a local patch, the estimated transmission becomes $\tilde{t}(\mathbf{x}) = \min_{y\in\Omega_x}\max_{z\in\Omega_y} t_b(z)$. The method employs a regularization formulation to obtain more robust values of the transmission for the entire image.

**Tang** ***et al.*** **2014** [43], unlike the previous methods, introduces a learning-based method. The method gathers multiscale features such as, dark channel [33], local maximum contrast [15], hue disparity, and local maximum saturation, and uses the random forest regressor [44] to learn the correlation between the features and the transmission, $t(\mathbf{x})$. The features related to the transmission are defined as follows:

$$\begin{aligned} F_D(\mathbf{x}) &= \min_{y\in\Omega(\mathbf{x})} \min_{c\in R,G,B} \frac{I^c(y)}{A^c}, \\ F_C(\mathbf{x}) &= \max_{y\in\Omega(\mathbf{x})} \sqrt{\frac{1}{3|\Omega(y)|}\sum_{z\in\Omega(y)} ||\mathbf{I}(y)-\mathbf{I}(z)||^2}, \\ F_H(\mathbf{x}) &= |\mathrm{H}(\mathbf{I}_{si}(\mathbf{x})) - \mathrm{H}(\mathbf{I}(\mathbf{x}))|, \\ F_S(\mathbf{x}) &= \max_{y\in\Omega(\mathbf{x})} \left(1 - \frac{\min_c I^c(y)}{\max_c I^c(y)}\right), \end{aligned} \tag{72}$$

where $I_{si} = \max[I^c(\mathbf{x}), 1 - I^c(\mathbf{x})]$. For the learning process, synthetic patches are generated from given haze-free patches, fixed white atmospheric light, and random transmission values, where the haze-free images are taken from the Internet. The paper claims that the most significant feature is the dark channel feature, however, other features also play important roles, particularly when the color of an object is the same that of the atmospheric light.

**Fattal 2014** [1] introduces another approach based on color lines. This method assumes that small image patches (*e.g.* 7×7) have a uniformly colored surface and the same depth, yet different shading. Hence, the model in Eq.(9) can be written as:

$$\mathbf{I}(\mathbf{x}) = l(\mathbf{x})\hat{\mathbf{R}} + (1-t)\mathbf{L}_\infty, \tag{73}$$

where $l(\mathbf{x})$ is the shading, and $\mathbf{R}(\mathbf{x}) = l(\mathbf{x})\hat{\mathbf{R}}$. Since the equation is a linear equation, in the RGB space the pixels of a patch will form a straight line (unless when the assumptions are violated, *e.g.* when patches containing color or depth boundaries). This line will intersect with another line formed by $(1-t)\mathbf{L}_\infty$. Since $\mathbf{L}_\infty$ is assumed to be known, then by having the intersection, $(1-t)$ can be obtained. To obtain $t(\mathbf{x})$ for the entire image, the method has to scan the pixels, extract patches, and find the intersections. Some patches might not give correct intersections, however if the majority of patches do, then the estimation can be correct. Patches containing object color identical to the atmospheric light color will not give any intersection, as the lines will be parallel. A Gaussian Markov random field (GMRF) is used to do the interpolation.

Sulami *et al.*'s method [45] uses the same idea and assumptions of the local color lines to estimate the atmospheric light, $\mathbf{L}_\infty$, automatically. First, it estimates the color of the atmospheric light by using a few patches, minimal two patches of different scene reflections. It assumes the two patches provide two different straight lines in the RGB space, and the atmospheric light's vector which starts from the origin must intersects with the two straight lines. Second, knowing the normalized color vector, it tries to estimate the magnitude of the atmospheric light. The idea is to dehaze the image using the estimated normalized light vector, then to minimize the distance between the estimated shading and the estimated transmission for the top 1 % brightness value found at each transmission level.

**Cai** *et al.* **2016** [46] proposes a learning based framework similar to [43] that trains a regressor to predict the transmission value $t(\mathbf{x})$ at each pixel ($16\times16$) from its surrounding patch.

TABLE III
SINGLE IMAGE DEHAZING METHODS WE COMPARED. PROGRAMMING LANGUAGE-M:MATLAB,P:PYTHON,C:C/C++. THE AVERAGE RUNTIME IS TESTED ON IMAGES OF RESOLUTION $720\times480$ USING A DESKTOP WITH XEON E5 3.5GHZ CPU AND 16GB RAM. WE USE THE CODE FROM THE AUTHORS, EXCEPT THOSE WITH MARKER *:WE IMPLEMENTED THE METHODS BY OURSELVES, †: WE DIRECTLY USE THE RESULTS FROM THE AUTHOR.

| Methods | Pub. venue | Code | Runtime($s$) |
|---|---|---|---|
| **Tarel 09** [35] | ICCV 2009 | M | 12.8 |
| **Ancuti 13** [40] | TIP 2013 | M* | 3.0 |
| **Tan 08** [15] | CVPR 2008 | C | 3.3 |
| **Fattal 08** [16] | ToG 2008 | M† | 141.1 |
| **He 09** [17] | CVPR 2009 | M* | 20 |
| **Kratz 09** [36] | ICCV 2009 | P | 124.2 |
| **Meng 13** [23] | ICCV 2013 | M | 1.0 |
| **Fattal 14** [1] | ToG 2014 | C† | 1.9 |
| **Berman 16** [18] | CVPR 2016 | M | 1.8 |
| **Tang 14** [43] | CVPR 2014 | M* | 10.4 |
| **Cai 16** [46] | arXiv 2016 | M* | 1.7 |

Unlike [43] that used hand crafted feature, Cai *et al.* [46] applied a convolutional neural network (CNN) based architecture with special network design (See [46] for the architecture). The network, termed DehazeNet, are conceptually formed by four sequential operations (feature extraction, multi-scale mapping, local extremum and non-linear regression), that consists of 3 convolution layers, a max-pooling, a Maxout unit and a Bilateral Rectified Linear Unit (BReLU, a nonlinear activation function extended from standard ReLU [47]). The training set used is similar to that in [43], namely they gathered haze free patches from Internet to generate hazy patches using the hazy imaging model with random transmissions $t$ and assuming white atmosphere light color ($\mathbf{L}_\infty = [1\,1\,1]^\top$). Once all the weights in the network are obtained from the training, the transmission estimation for a new hazy image patch is simply forward propagation using the network. To handle the block artifact caused by the patch based estimation, guided filtering [48] is used to refine the transmission map.

**Berman** *et al.* **2016** [18] proposes an algorithm based on a new, non-local prior. This is a departure from existing methods (*e.g.* [15], [17], [23], [1], [43], [46] *etc.*) that use patch based transmission estimation. The algorithm by [18] relies on the assumption that colors of a haze-free image are well approximated by a few hundred distinct colors, that form tight clusters in RGB space and pixels in a cluster are often non-local (spread in the whole images). The presence of haze will elongate the shape of each cluster to a line in color space as the pixels may be affected by different transmission coefficients due to their different distances to the camera. The lines, termed haze-line, is informative in estimating the transmission factors. In their algorithm, they first proposed a clustering method to group the pixels and each cluster becomes a haze-line. Then the maximum radius of each cluster is calculated and used to estimate the transmission. A final regulation step is performed to enforce the smoothness of the transmission map.

## IV. QUANTITATIVE BENCHMARKING

In this section, we benchmark some visibility enhancement methods. Our focus is on recent single-image based methods. Compared with other approaches, single-image based approach is more practical and thus have more potential applications. By benchmarking the methods in this approach, we consider it will be beneficial, since one can know the comparisons of the methods quantitatively.

To compare all methods quantitatively we need to test on dataset with ground truth. Ideally, similar to what Narasimhan *et al.* [49] had done, the dataset should be created from real atmospheric scenes taken over a long period of time to have all possible atmospheric conditions ranging from light mist to dense fog with various backgrounds of scenes. While it may be possible, it is not trivial at all, since it has to be done in certain time and locations where fog and haze are present frequently and the scene, the illumination should keep fixed (which means clouds and sunlight distribution should be about the same). Unfortunately, these conditions rarely meet. Moreover, it is challenging to have a pixel-wise ground truth of a scene without the effect of particles even in a clear day, particularly for distant objects, as significant amount of atmospheric particles are always present. These reasons motivated us to use synthesised data. We first perform dehazing evaluations on a recent dataset provided by Fattal [1]. Moreover we create another dataset with physics based rendering technique for the evaluation. In the following sections, we will describe the details of the dataset and present the results of different dehazing methods on these datasets.

We compare 11 dehazing methods in total including most representative dehazing methods published in major venues, as listed in Table III. We use the code from the authors for evaluation if the source codes are available. We also implement [40], [17], [43], [46] by strictly following the pipeline and parameter settings described in the paper. For [16] and [1], we directly use the results provided along the dataset [1]. Following the convention in the dehazing papers, we simply use the first author's name with year of publication (*e.g.* **Tan 08**) to indicate each method.

### A. Evaluation on Fattal's Dataset [1]

Fattal's dataset [1] [2] has 11 haze images generated using real images with known depth maps. Assuming a spatially constant scattering coefficient $\beta$, the transmission map can be generated by applying the direct attenuation model, and the synthesized haze image can be generated using the haze model Eq. (12). One example of the synthesized images is shown in Figure 5.

As mentioned in Sec 2.3, there are generally three major steps in dehazing: (1) estimation of the atmospheric light, (2) the estimation of the transmission (or the airlight), and (3) the final image enhancement that imposes a smooth constraint of the neighboring transmission. A study of the atmospheric light color estimation in dehazing can be found in [45]. In our benchmarking, our focus is on evaluating the transmission map estimation and final dehazing results. We therefore directly use ground truth atmospheric light color provided in the dataset for all dehazing methods.

***Transmission map evaluation*** Table IV lists the mean absolute difference (MAD) of the estimated transmissions (excluding sky regions) of each method to the ground truth transmission. Note that, two methods, **Tarel 09** [35] and **Ancuti 13** [40], are not included, as they do not require the transmission estimation. The three smallest errors for each image are highlighted. As can be seen, there is no one method can be outstanding for all cases. The recent method **Fattal 14** [1] and **Berman 16** [18] can obtain more accurate estimation of the transmission for most cases. The early work of **Tan 08** [15] gives less precise estimation. Another early work **Fattal 08** [16] is not stable and it obtains accurate estimation on a few cases (*e.g. flower2, reindeer*) while obtains largest error on some other cases (*e.g. church, road1*).

We plot the average MAD over all 11 cases in Figure 4. It is noticed that in general, latest methods perform better in the transmission estimation. The method of **Fattal 14** [1] and **Berman 16** [18] rank at the top place, while the two learning based method **Tang 14** [43] and **Cai 16** [46] are at the second place. However, we noticed in our experiments that the learning based methods heavily rely on the white balance step with correct atmospheric light color. Once there are small errors in atmospheric light color estimation, their performance drops quickly. This indicates the learned models are actually overfilled to the case of white balanced haze images as in the training process it always assume pure white atmosphere light color. **He 09** [17]'s results also are at a decent rank place. This demonstrates that dark channel prior is an effective prior in the transmission estimation.

We further test the mean signed difference (MSD) on the transmission estimation results (excluding sky regions) as $\text{MSD} = \frac{1}{N}\sum_i(\tilde{t}_i - t_i)$, where $i$ is the pixel index, $N$ is the total number of pixels, $\tilde{t}$ is the estimated transmission, and $t$ is the ground truth transmission. By doing so, we can test whether a method overestimates (positive signed difference) or underestimates (negative signed difference) the transmission, which can not be revealed using the previous MAD metrics. The MSDs are listed in Table V and the average MSDs are plotted in Figure 4. It is observed that **Tan 08** [15] mostly underestimate the transmission, as a result it obtains over-saturated dehaze results. **Fattal 08** [16], on the contrary, likely overestimate the transmission, leading to a results with haze still presented in the output. The two methods **He 09** [17] and **Meng 13** [23] also slightly underestimate the transmission due to the fact they essentially predict the lower bound of transmission.

***Dehazing results evaluation*** We evaluate the dehazing results. The mean absolute difference (MAD) of each method (excluding sky regions) to the ground truth clean image are listed in Table VI and the dehazing results on $church$ case are shown in Figure 5. In Table VI, the three smallest errors for each image are highlighted. Again, there is no one method can be outstanding for all cases. It is clear that the observed non-model based methods **Tarel 09** [35] and **Ancuti 13** [40] obtain largest error in the recovery. The visual qualities of their results are also rather inferior compared with other methods (as can be seen in Figure 5). This shows that only image contrast enhancement operation without the haze image model Eq. (12) cannot achieve satisfactory results. Among the rest model based methods, the latest methods **Meng 13** [23], **Tang 14** [43], **Fattal 14** [1], **Cai 16** [46], **Berman 16** [18], and also **He 09** [17] generally perform better than early dehazing methods **Tan 08** [15], **Fattal 08** [16], and **Kratz 09** [36]. Especially **Fattal 14** [1] and **Berman 16** [18] are *the best two methods* that can provide dehazing results that are the most close to the ground truth. This quantification ranking reflects well the visual quality as the example shown in Fig 5.

[2]http://www.cs.huji.ac.il/~raananf/projects/dehaze_cl/results/index_comp.html

***Evaluation with various haze levels*** Additionally, we test the performance of each method for different haze levels. In Fattal's dataset [1], he provides a subset of images (*lawn1, mansion, reindeer, road1*) that are synthesized with three different haze level by control the scattering coefficient $\beta$. As $\beta$ increases, denser haze effects will appear. We measure the transmission estimation error and final dehazing error using the mean absolute difference and the average results over all scenes are plotted in Figure 6.

It is clearly observed that **Fattal 14** [1] stably stand out in achieving less errors in both transmission estimation and final dehazing at different haze levels. **Fattal 08** [16] works well only at low haze levels and the performance drops at medium and high haze levels. Looking at the transmission results, we can see **Tan 08** [15], **He 09** [17], and **Meng 13** [23]'s estimation becomes more accurate when haze level increases. This demonstrates the priors of these three methods are correlated with haze such that these priors can tell more information with more haze. The difference is that **He 09** [17], and **Meng 13** [23] can achieve much smaller transmission errors than **Tan 08** [15], showing the superiority of dark channel prior [17] and boundary constrain [23] against the local contrast [15] for this task. This can be explained by the fact that with heavier haze, the contribution of the airlight $A(x)$ increases, where dark channel prior and boundary constraint assumption can fit more.

**Berman 16** [18] can achieve the least transmission estimation error at medium haze level but the error increases at both low and heavy haze levels. This may reveal one limitation of **Berman 16** [18] that the haze lines formed from non local pixels work well only at certain haze levels. In near clean (low haze level) or heavily hazy scenarios, the haze lines found may not be reliable. The two learning methods, **Tang 14** [43] and **Cai 16** [46], predict the transmission decently well. For the final dehaze results, most methods obtain large error in heavy haze except **He 09** [17] and **Fattal 14** [1].

### B. Evaluation on our Dataset

Unlike the Fattal's dataset that are generated using image with the haze image model Eq. (12), we generate our dataset using a physically based rendering technique (PBRT) that uses the Monte Carlo ray tracing in a volumetric scattering medium [50]. We render five sets of different scenes under different haze level of different types, namely *swarp, house, building, island, villa*. Our scenes are created using freely available 3D models. All five scenes contains large depth variation from a few meters to about $2,000$ meters. We assume a uniform haze density in the space and use homogeneous volumes in our rendering. For each of the five scenes, we render six images. The first one is rendered with no participating media and is considered as the ground truth. The remaining five images are rendered with increasing haze level, namely, by evenly increasing the absorption coefficient $\sigma_a$ and the scattering coefficient $\sigma_s$. Figure 7 shows two sets of our generated synthetic data (*building, island*). As can be seen, the visibility of the scene, especially further away objects, decreases when the haze level increases. The whole dataset will be available via a project website.

We have evaluated 9 methods on our dataset (**Fattal 08** [16] and **Fattal 14** [1]'s results are not available on our dataset). We quantify the visibility enhancement outputs by comparing them with their respective ground truths. The quantitative measurement is done by using the structural similarity index (SSIM) [51]. While MAD directly measure the closeness of the output to the ground truth pixel by pixel, SSIM is more consistent with human visual perception, especially in the cases of dehazing for heavier haze level. SSIM is a popular choice to compute the structure similarity of two images in image restoration tasks. Unlike MAD, a higher value in SSIM indicate a better match as it is a similarity measurement.

Figure 8 shows the performance of each method in term of SSIM. It is observed that again latest methods **Tang 14** [43], **Cai 16** [46], and **Berman 16** [18] generally performed better than others. **He 09** [17] also performs very well, especially in heavier haze levels. This is consistent with our experiment in Section IV-A.

### C. Qualitative results on real images

We have also list three qualitative examples of the dehazing results on real hazy images by different methods (more visual comparisons can be found in the previous dehazing paper *e.g.* [1], [18]). The visual comparison here confirms our findings in the previous benchmarking that **Fattal 14** [1] and **Berman 16** [18] are *the best two methods* that can consistently provide excellent dehazing results. Some early methods like **Kratz 09** [36], **Tarel 09** [35], **Ancuti 13** [40] exhibit noticeable limitations in the dehazing results (*e.g.* over-saturation, boundary artifacts, color shift). The **He 09** [17] and **Meng 13** [23] also performs well and obtain similar results as they essentially both predict the lower bound of the transmission. The results of the learning based method **Tang 14** [43] and **Cai 16** [46] can produce appealing results in general but usually tend to leave haze in results.

## V. Summary and Discussion

**Summary** This paper has provided a thorough survey of major methods of visibility enhancement in hazy/foggy scenes. Various modalities such as multiple images, known approximated depth, stereo, and polarizing filters have been introduced to tackle the problem. Special emphasis was placed on single-image methods where significant image cues have been explored to enhance visibility, such as local contrast [15], shading-transmission decomposition [16], dark channel prior [17], line intersection [1] *etc.*. The tenet of all the methods is to use scene cues to estimate light transmission and to unveil scene reflection based on the estimated transmission.

TABLE IV
THE MEAN ABSOLUTE DIFFERENCE OF TRANSMISSION ESTIMATION RESULTS ON FATTAL'S DATASET [1]. THE THREE SMALLEST VALUES ARE HIGHLIGHTED.

| Methods | Church | Couch | Flower1 | Flower2 | Lawn1 | Lawn2 | Mansion | Moebius | Reindeer | Road1 | Road2 |
|---|---|---|---|---|---|---|---|---|---|---|---|
| **Tan 08** [15] | 0.167 | 0.367 | 0.216 | 0.294 | 0.275 | 0.281 | 0.316 | 0.219 | 0.372 | 0.257 | 0.186 |
| **Fattal 08** [16] | 0.377 | 0.090 | 0.089 | 0.075 | 0.317 | 0.323 | 0.147 | 0.111 | 0.070 | 0.319 | 0.347 |
| **Kratz 09** [36] | 0.147 | 0.096 | 0.245 | 0.275 | 0.089 | 0.093 | 0.146 | 0.239 | 0.142 | 0.120 | 0.118 |
| **He 09** [17] | 0.052 | 0.063 | 0.164 | 0.181 | 0.105 | 0.103 | 0.061 | 0.208 | 0.115 | 0.092 | 0.079 |
| **Meng 13** [23] | 0.113 | 0.096 | 0.261 | 0.268 | 0.140 | 0.131 | 0.118 | 0.228 | 0.128 | 0.114 | 0.096 |
| **Tang 14** [43] | 0.141 | 0.074 | 0.044 | 0.055 | 0.118 | 0.127 | 0.096 | 0.070 | 0.097 | 0.143 | 0.158 |
| **Fattal 14** [1] | 0.038 | 0.090 | 0.047 | 0.042 | 0.078 | 0.064 | 0.043 | 0.145 | 0.066 | 0.069 | 0.060 |
| **Cai 16** [46] | 0.061 | 0.114 | 0.112 | 0.126 | 0.097 | 0.102 | 0.072 | 0.096 | 0.095 | 0.092 | 0.088 |
| **Berman 16** [18] | 0.047 | 0.051 | 0.061 | 0.115 | 0.032 | 0.041 | 0.080 | 0.153 | 0.089 | 0.058 | 0.062 |

TABLE V
THE MEAN SIGNED DIFFERENCE OF TRANSMISSION ESTIMATION RESULTS ON FATTAL'S DATASET [1].

| Methods | Church | Couch | Flower1 | Flower2 | Lawn1 | Lawn2 | Mansion | Moebius | Reindeer | Road1 | Road2 |
|---|---|---|---|---|---|---|---|---|---|---|---|
| **Tan 08** [15] | 0.013 | -0.339 | -0.117 | -0.268 | -00.083 | -0.089 | -0.301 | -0.160 | -0.358 | -0.148 | -0.117 |
| **Fattal 08** [16] | 0.376 | 0.088 | 0.088 | 0.071 | 0.317 | 0.323 | 0.143 | 0.073 | 0.063 | 0.312 | 0.327 |
| **Kratz 09** [36] | -0.006 | 0.010 | -0.220 | -0.267 | 0.003 | -0.013 | -0.114 | -0.236 | -0.083 | -0.030 | 0.067 |
| **He 09** [17] | -0.035 | -0.045 | -0.162 | -0.180 | -0.091 | -0.086 | -0.041 | -0.208 | -0.105 | -0.054 | -0.047 |
| **Meng 13** [23] | -0.112 | -0.003 | -0.259 | -0.266 | -0.139 | -0.130 | -0.101 | -0.223 | -0.086 | -0.109 | -0.089 |
| **Tang 14** [43] | 0.133 | 0.054 | -0.008 | -0.046 | 0.059 | 0.067 | 0.089 | -0.051 | 0.013 | 0.094 | 0.123 |
| **Fattal 14** [1] | -0.019 | 0.086 | -0.021 | -0.019 | 0.063 | 0.045 | 0.002 | -0.105 | 0.006 | 0.005 | -0.015 |
| **Cai 16** [46] | -0.002 | 0.086 | -0.096 | -0.118 | 0.012 | 0.017 | -0.028 | -0.070 | 0.044 | 0.001 | 0.023 |
| **Berman 16** [18] | 0.009 | -0.014 | -0.051 | -0.115 | -0.008 | -0.013 | -0.076 | -0.152 | -0.059 | -0.041 | -0.021 |

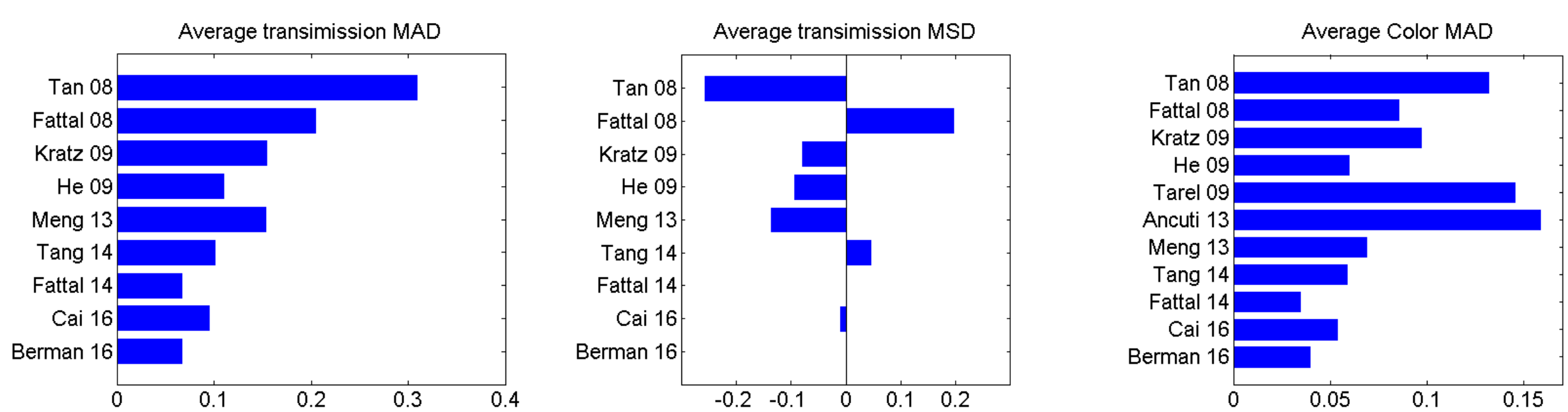

Fig. 4. The average performance of different dehazing methods on Fattal's dataset [1].

Furthermore, there are two principle properties of the transmission estimation: the estimation of the atmospheric light (both its color and intensity) and the smoothness constraint of the transmission.

We have also conducted the first quantitative benchmark for most representative single image dehazing methods. Our primary finding from the benchmark is that recent works [1],[18] generally performs better in the dehazing tasks. Machine learning based methods [43], [46] can also get decent results, but their performance are likely to be affected by the white balancing step. Therefore we still recommend the prior based methods [1],[18] over the learning based methods [43], [46] in practical use for robustness. We also found that the popular dark channel prior [17] is an effective prior in dehazing, especially for denser haze levels.

For the dataset used in the benchmark, we picked a dataset from Fattal [1] and also our newly introduced synthetic dataset which provides ground truth clean image and haze images with different haze levels. We hope the community can take benefit of our dataset by being able to assess new methods more objectively and to find new problems in the field.

**Discussion** When fog is considerably thick, the problem of visibility enhancement becomes harder. This is because scene reflection is "buried" further underneath the airlight ($\mathbf{A}$) and transmission ($t$). Considering the scattering model in Eq.(9), when the scattering coefficient $\beta$ is large, *i.e.* in a thick foggy scene, the transmission ($t = e^{-\beta d}$) is small. Consequently, the airlight ($\mathbf{A} = (1-t)\mathbf{L}_\infty$) is dominated by the atmospheric light, $\mathbf{L}_\infty$, and thus the veiling component takes up more portion in the image intensity. Also, since the transmission is small, the contribution of scene reflection in the image intensity becomes reduced significantly, due to the multiplication of $\mathbf{R}$ with a fractionally small value of $t$. The combined airlight and transmission components hide the underlying scene reflection information in the image intensities.

Based on this, some questions might arise: how do we know whether the information of scene reflection is too minuscule to be recovered? How thick is fog that we cannot extract the scene reflection any longer? Answering such questions are important theoretically, since then we can know the limit of visibility enhancement in bad weather. Furthermore, in thick foggy scenes, due to absorption and scattering to directions other than the line of sight, image blur will be present more prominently and it is not modeled in the two flux scattering model.

TABLE VI
THE MEAN ABSOLUTE DIFFERENCE OF FINAL DEHAZING RESULTS ON FATTAL'S DATASET [1]. THE THREE SMALLEST VALUES ARE HIGHLIGHTED.

| Methods | Church | Couch | Flower1 | Flower2 | Lawn1 | Lawn2 | Mansion | Moebius | Reindeer | Road1 | Road2 |
|---|---|---|---|---|---|---|---|---|---|---|---|
| **Tan 08** [15] | 0.109 | 0.139 | 0.098 | 0.134 | 0.146 | 0.146 | 0.154 | 0.131 | 0.150 | 0.111 | 0.139 |
| **Fattal 08** [16] | 0.158 | 0.055 | 0.028 | 0.022 | 0.116 | 0.123 | 0.071 | 0.039 | 0.034 | 0.135 | 0.165 |
| **Kratz 09** [36] | 0.099 | 0.060 | 0.155 | 0.161 | 0.055 | 0.059 | 0.085 | 0.155 | 0.083 | 0.073 | 0.088 |
| **He 09** [17] | 0.036 | 0.038 | 0.078 | 0.080 | 0.056 | 0.057 | 0.034 | 0.121 | 0.061 | 0.051 | 0.052 |
| **Tarel 09** [35] | 0.173 | 0.112 | 0.130 | 0.120 | 0.146 | 0.161 | 0.113 | 0.143 | 0.179 | 0.148 | 0.176 |
| **Ancuti 13** [40] | 0.188 | 0.078 | 0.276 | 0.219 | 0.128 | 0.144 | 0.109 | 0.189 | 0.145 | 0.135 | 0.142 |
| **Meng 13** [23] | 0.052 | 0.060 | 0.114 | 0.106 | 0.055 | 0.055 | 0.048 | 0.096 | 0.065 | 0.052 | 0.054 |
| **Tang 14** [43] | 0.087 | 0.048 | 0.017 | 0.019 | 0.072 | 0.078 | 0.053 | 0.031 | 0.053 | 0.088 | 0.106 |
| **Fattal 14** [1] | 0.025 | 0.053 | 0.019 | 0.015 | 0.035 | 0.033 | 0.022 | 0.076 | 0.034 | 0.033 | 0.038 |
| **Cai 16** [46] | 0.042 | 0.069 | 0.045 | 0.049 | 0.061 | 0.0652 | 0.040 | 0.043 | 0.053 | 0.057 | 0.065 |
| **Berman 16** [18] | 0.032 | 0.031 | 0.022 | 0.045 | 0.026 | 0.031 | 0.049 | 0.081 | 0.045 | 0.040 | 0.042 |

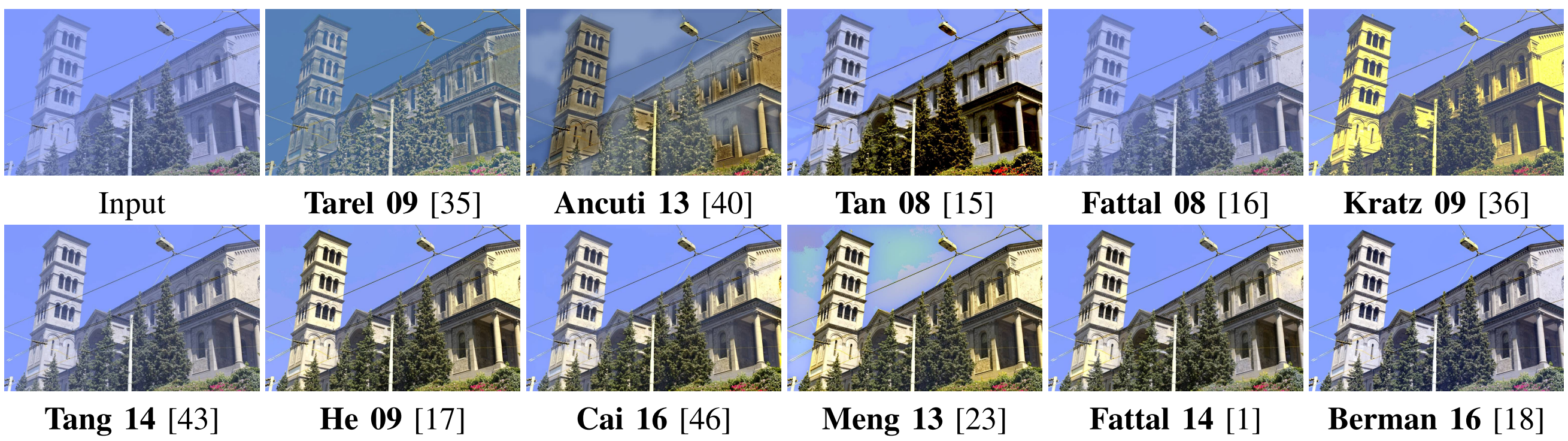

Fig. 5. Final haze removal results on the *church* case.

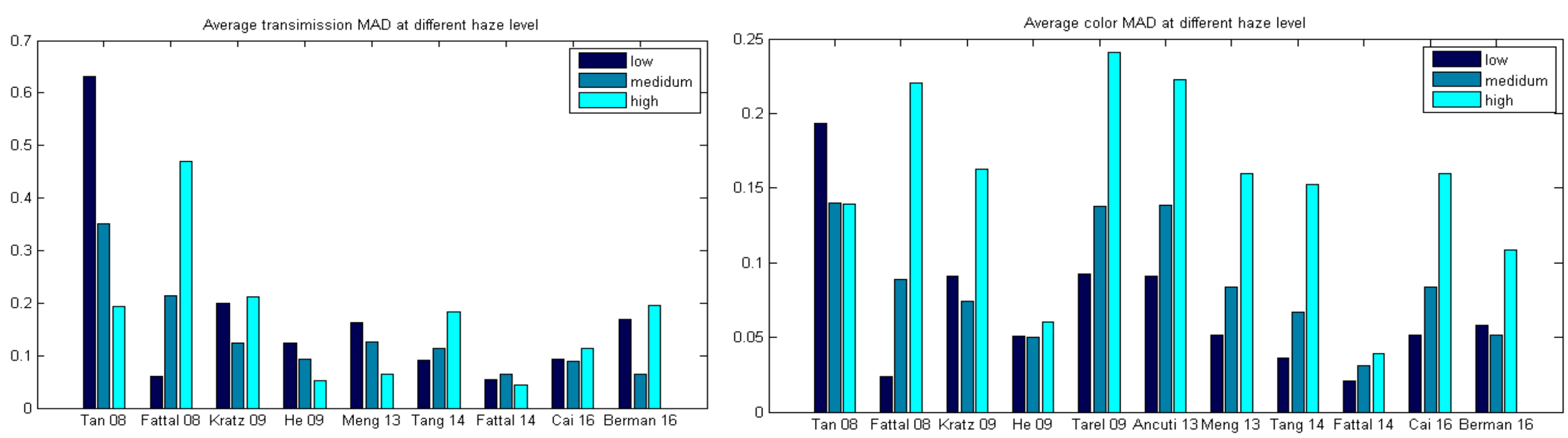

Fig. 6. Comparisons of the results for different haze levels.

Another issue to note here is the gamma correction or tone-curve applied to most images captured by cameras. Although many methods do not explicitly mention the assumption of linearity between the flux of incoming light and the pixel intensity values, based on the scattering model (Eq.9), there is a clear linear model. This that the correlation between the image intensity (**I**) and the remaining parameters is not truly linear as described in the model due to the non-linear intensity manipulation applied onboard the cameras. While for the purpose of visibility enhancement this might not be an issue, for physically correct scene reflection recovery, proper gamma correction and camera color transformation might need to consider.

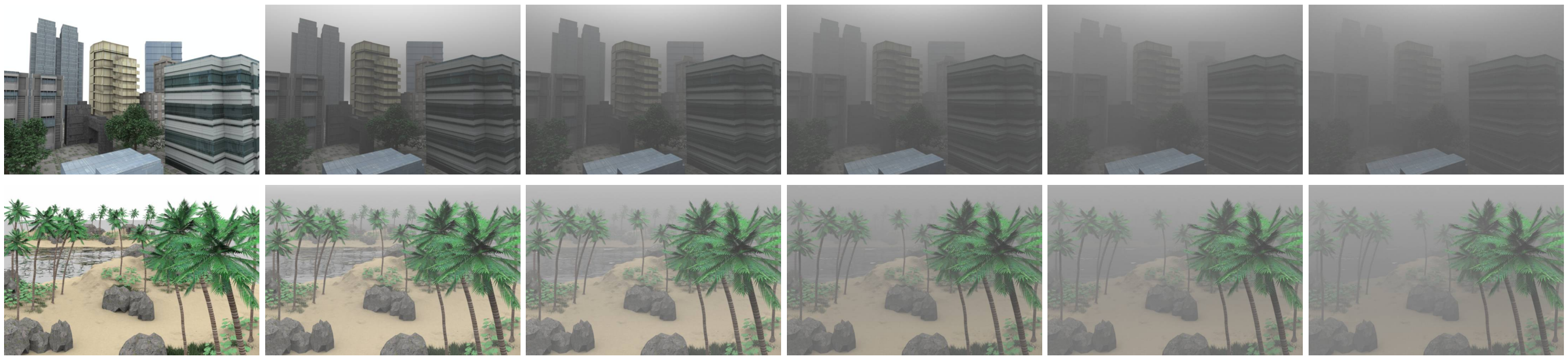

Fig. 7. Samples of our synthetic data with increasing haze levels.

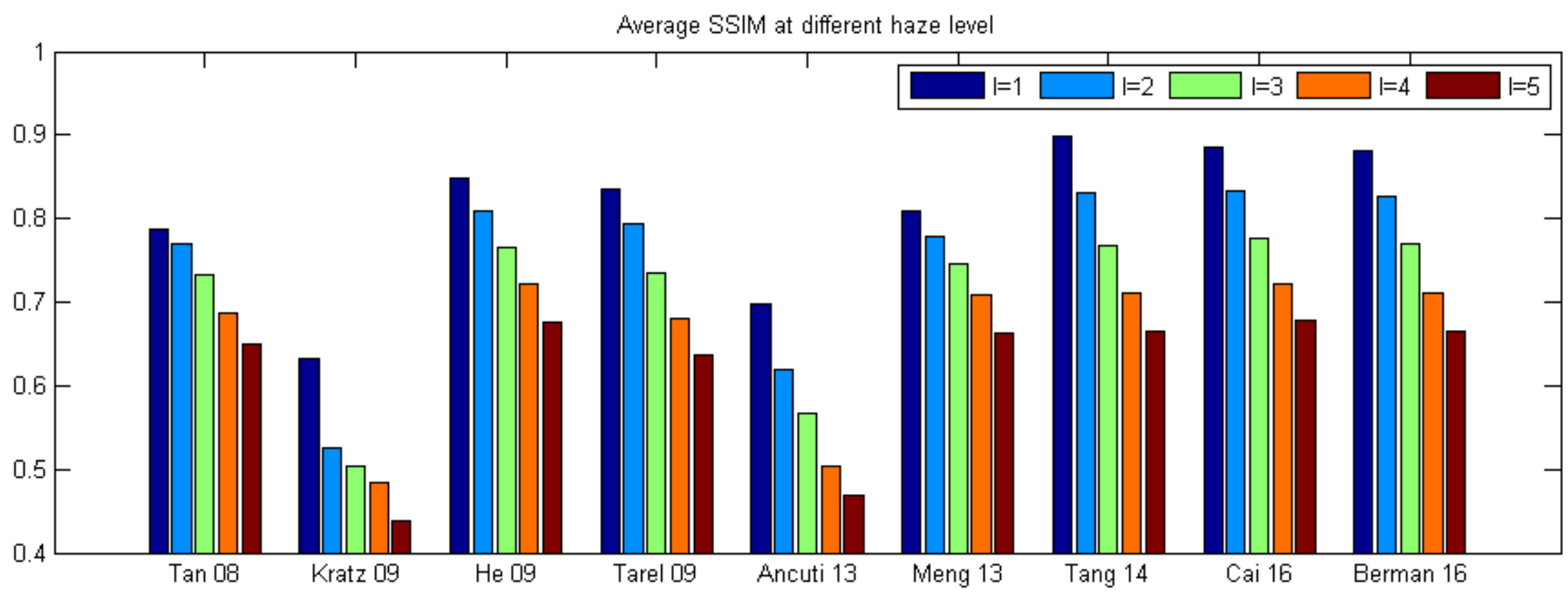

Fig. 8. The performance of each method on our dataset on 5 haze levels (l=1,2,3,4,5, low to high) in term of SSIM.

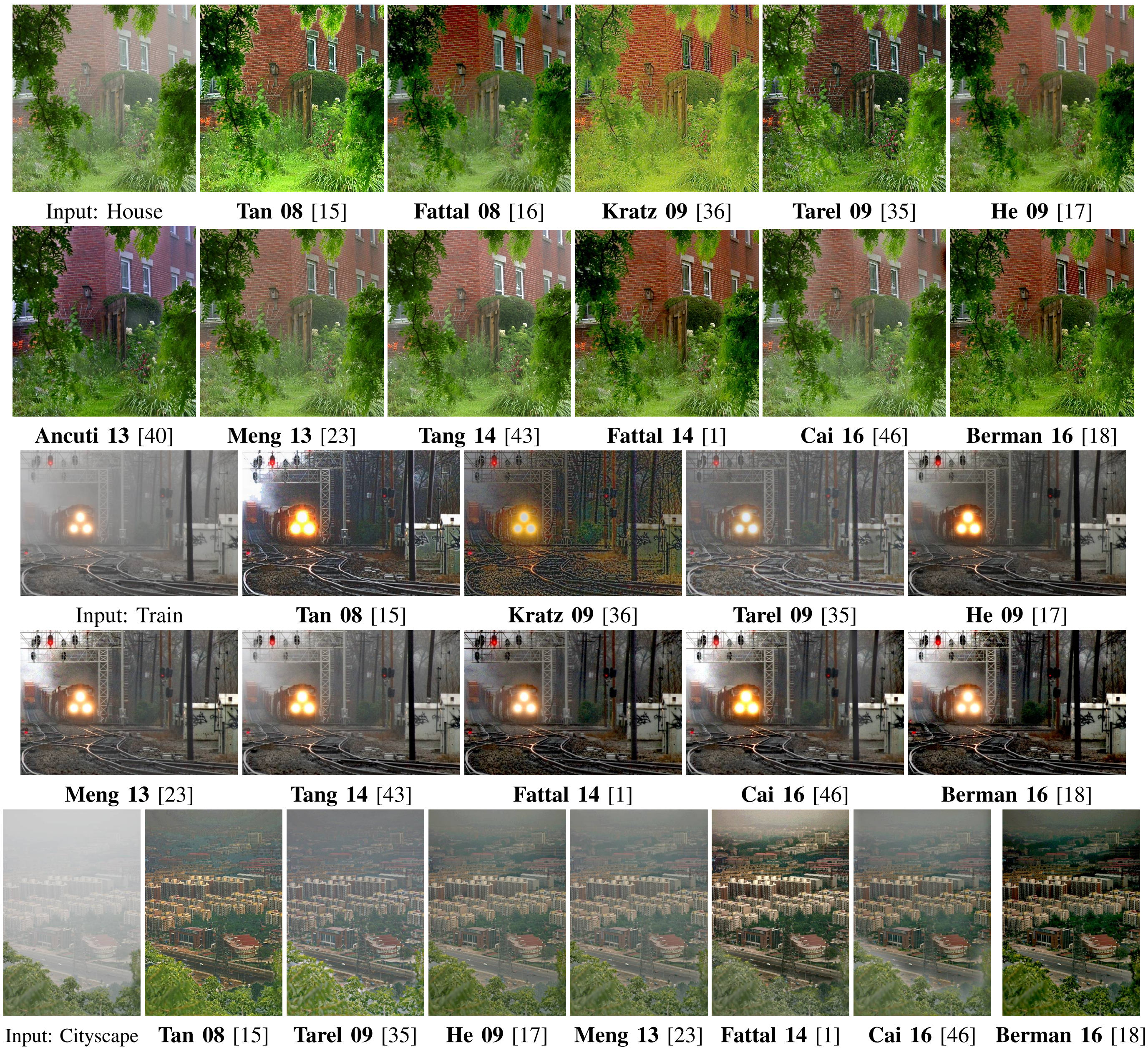

Fig. 9. Example comparison on real images.